\RequirePackage{fix-cm}
\documentclass[12pt, a4paper, twoside, notitlepage]{article}
\usepackage[a4paper]{geometry}
\newgeometry{left=2.5cm,right=2.5cm,bottom=2cm,top=1cm}
\usepackage[utf8]{inputenc}
\usepackage[T1]{fontenc}
\usepackage{lmodern} 

\usepackage{times}%

\usepackage{amsmath}
\usepackage{amsfonts}
\usepackage{dcolumn}
\usepackage{endnotes}
\usepackage{graphics}
\usepackage{graphicx}
\usepackage{caption}
\usepackage{subcaption}
\usepackage{multirow}

\usepackage[table]{xcolor}
\definecolor{lightgray}{gray}{0.96}

\usepackage{marvosym}
\usepackage[gen]{eurosym}

\DeclareUnicodeCharacter{20AC}{\euro{}}

\usepackage{acronym}
\acrodef{DER}{Distributed Energy Resources}
\acrodef{MTBF}{Mean Time Between Failures}
\acrodef{MTTR}{Mean Time To Recover}

\usepackage{tikz}
\usetikzlibrary{shadows,calc}
\def\shadowshift{3pt,-3pt}
\def\shadowradius{6pt}

\colorlet{innercolor}{black!60}
\colorlet{outercolor}{gray!05}

\newcommand\drawshadow[1]{
    \begin{pgfonlayer}{shadow}
        \shade[outercolor,inner color=innercolor,outer color=outercolor] ($(#1.south west)+(\shadowshift)+(\shadowradius/2,\shadowradius/2)$) circle (\shadowradius);
        \shade[outercolor,inner color=innercolor,outer color=outercolor] ($(#1.north west)+(\shadowshift)+(\shadowradius/2,-\shadowradius/2)$) circle (\shadowradius);
        \shade[outercolor,inner color=innercolor,outer color=outercolor] ($(#1.south east)+(\shadowshift)+(-\shadowradius/2,\shadowradius/2)$) circle (\shadowradius);
        \shade[outercolor,inner color=innercolor,outer color=outercolor] ($(#1.north east)+(\shadowshift)+(-\shadowradius/2,-\shadowradius/2)$) circle (\shadowradius);
        \shade[top color=innercolor,bottom color=outercolor] ($(#1.south west)+(\shadowshift)+(\shadowradius/2,-\shadowradius/2)$) rectangle ($(#1.south east)+(\shadowshift)+(-\shadowradius/2,\shadowradius/2)$);
        \shade[left color=innercolor,right color=outercolor] ($(#1.south east)+(\shadowshift)+(-\shadowradius/2,\shadowradius/2)$) rectangle ($(#1.north east)+(\shadowshift)+(\shadowradius/2,-\shadowradius/2)$);
        \shade[bottom color=innercolor,top color=outercolor] ($(#1.north west)+(\shadowshift)+(\shadowradius/2,-\shadowradius/2)$) rectangle ($(#1.north east)+(\shadowshift)+(-\shadowradius/2,\shadowradius/2)$);
        \shade[outercolor,right color=innercolor,left color=outercolor] ($(#1.south west)+(\shadowshift)+(-\shadowradius/2,\shadowradius/2)$) rectangle ($(#1.north west)+(\shadowshift)+(\shadowradius/2,-\shadowradius/2)$);
        \filldraw ($(#1.south west)+(\shadowshift)+(\shadowradius/2,\shadowradius/2)$) rectangle ($(#1.north east)+(\shadowshift)-(\shadowradius/2,\shadowradius/2)$);
    \end{pgfonlayer}
}

\pgfdeclarelayer{shadow}
\pgfsetlayers{shadow,main}

\newsavebox\mybox
\newlength\mylen

\newcommand\shadowimage[2][]{%
\setbox0=\hbox{\includegraphics[#1]{#2}}
\setlength\mylen{\wd0}
\ifnum\mylen<\ht0
\setlength\mylen{\ht0}
\fi
\divide \mylen by 120
\def\shadowshift{\mylen,-\mylen}
\def\shadowradius{\the\dimexpr\mylen+\mylen+\mylen\relax}
\begin{tikzpicture}
\node[anchor=south west,inner sep=0] (image) at (0,0) {\includegraphics[#1]{#2}};
\drawshadow{image}
\end{tikzpicture}}

\begin{document}

\title{Assisted Energy Management in Smart Microgrids}

 \author{Andrea Monacchi\\
	Wilfried Elmenreich\\
	Institute of Networked and Embedded Systems\\
   Alpen-Adria-Universität Klagenfurt\\}

\maketitle

\begin{abstract}
Demand response provides utilities with a mechanism to share with end users the stochasticity resulting from the use of renewable sources.
Pricing is accordingly used to reflect energy availability, to allocate such a limited resource to those loads that value it most.
However, the strictly competitive mechanism can result in service interruption in presence of competing demand.
To solve this issue we investigate on the use of forward contracts, i.e., service-level agreements priced to reflect the expectation of future supply and demand curves.
Given the limited resources of microgrids, service interruption is an opposite objective to the one of service availability.
We firstly design policy-based brokers and identify then a learning broker based on artificial neural networks.
We show the latter being progressively minimizing the reimbursement costs and maximizing the overall profit.
\end{abstract}

\section{Introduction}\label{sec:introduction}
Demand response has been advocated as potential solution to compensate the growing instability caused by energy generation from renewable sources and the current shift towards electric mobility.
A microgrid is defined as a small power system built from the aggregation of energy sources and small loads, and is able to operate as an independent power island when needed \cite{5275343}.
As such, microgrids provide a potential solution to the problem of remote electrification, where the connection to the main power grid is technically or economically unfeasible.
Consequently, microgrids might be employed for electrification of rural villages, campuses, smart buildings, such as residential and office spaces, and are typically composed of:
i) electrical loads, ii) energy storage elements to accumulate excessive energy and iii) a gateway running the controller that regulates the power flow between the local and the main power grids.
\begin{figure}[h!]
	\centering
	\includegraphics[trim=1.545cm 24.65cm 8.555cm 1.7cm,clip,width=0.8\columnwidth]{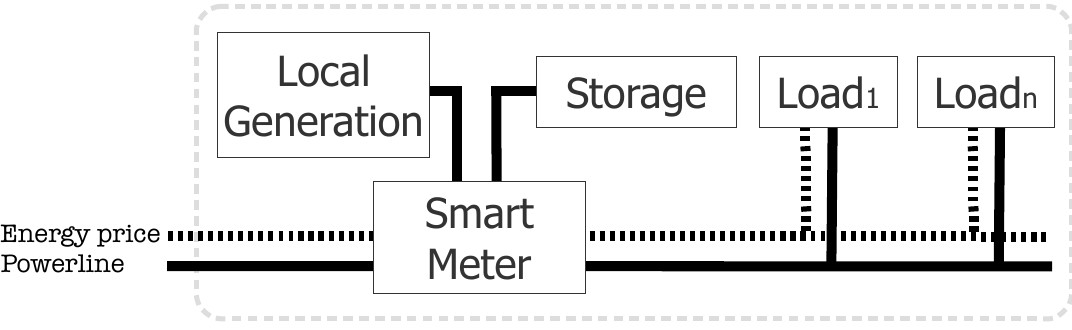}
	\caption{A smart microgrid}
	\label{fig:market}
\end{figure}
Management of energy resources in power grids is normally performed by means of electricity markets.
We can distinguish in i) wholesale electricity markets in which generators compete to supply their output to retailers
and ii) retail electricity markets in which end-use customers can select their supplier from a pool of competing retailers.
For final customers this consists in selecting the utility that provides the best provisioning tariff.
Energy trading between generators and retailers is a complex problem which can result in high price volatility.
This is due to the uncertainty resulting from the involved generation facilities (e.g., fuel costs and weather),
as well the difficulty of retailers to effectively predict consumers' demand for more than a few days in the future.
Forward contracts offer a way of protection towards price volatility, which in turn allows generators for better planning investments on generation plants.
In addition, the employment of demand response as a congestion pricing scheme can share such uncertainty also with end customers.
A price signal can be used to reflect the availability of the shared and scarce resource, to consequently allocate it to the users that value it most.
This in turn limits the need for adaptation at generation side, which would otherwise imply over-dimensioning of generation capability, the use of fossil fuels, or costly storage facilities.
In a microgrid, such a real-time market can coordinate the operation of loads, thus avoiding blackouts and contributing to flattening the demand (i.e., the peak-to-average ratio).
Loads can strategically decide their purchases, as well as curtail their demand and allocate a shorter duration than actually needed.
Moreover, the notion of money exchange provides an understandable means to communicate the results of the energy management system as well as easily capture users' preferences \cite{monacchi:2013Nov}.
However, requirements for microgrids are more stringent than for main power grids.
In larger grids, wholesale markets are commonly held for hourly intervals on a day-aheady basis, due to the physical limits for the actuation of generators.
While this offers a good solution to plan resources, it becomes a critical issue to timely react in presence of a relevant portion of renewable energy generation.
Moreover, the more limited amount of electricity and customers involved in microgrids requires the scheduling of power at a finer time resolution.
This is important to avoid underuse of resources. A state of an electrical appliance could demand provisioning over seconds rather than hours.
Sizing of the allocation interval reproduces the problems of internal and external fragmentation experienced in dynamic memory allocation \cite{Silberschatz:2008}.
Dividing a shared resource in predefined-size partitions limits the size of the process to be run and thus the degree of multiprogramming. 
Specifically, while a large allocation interval can be used to suit any load state this would often lead to underuse (i.e., internal fragmentation).
On the contrary, small allocation intervals could be used to timely react to system changes, although this would cause operation interruption.
In \cite{monacchi2014:hems} we implemented a market mechanism with a short allocation time of 1 second to highlight this problem.
The mechanism reallocates power at each trading cycle, which results in service interruptions with competing demand.
To mitigate such issues we investigate on the use of forward contracts in microgrids in this study.
We propose the use of a power broker, which is able to formulate prices based on his model of future demand and supply.
We consider a scenario with photovoltaic power generation as this was shown in \cite{monacchi:2013Nov} being the most employed renewable energy source in domestic settings.
The broker is able to formulate prices for multiple service-level agreements, each defining a different duration for power provisioning.
In this way, the multiple prices reflect differences in terms of generation uncertainty and planned congestion, thus offering different levels of quality of service.

The remainder of this paper is structured as follows:
a survey of previous work on market-based energy management opens the discussion in Sect.~\ref{sec:relwork}.
The brokerage problem is formalized as a cost minimization function in Sect.~\ref{sec:problem}.
We introduce service-level agreements to mitigate the uncertainty and service interruptions which motivate this study.
An experiment is set up to assess effects on system-wide performance.
In particular, Sect.~\ref{sec:experiments} reports the evaluation metrics and the modeled scenario.
The effect of provisioning agreements on the selected metrics is discussed in Sect.~\ref{sec:coordination}.
The following Sect.~\ref{sec:annevaluation} reports an extensive evaluation of candidate broker designs.
Finally, Sect.~\ref{sec:conclusions} summarizes our contribution and lists open questions to be further investigated in future.

\section{Related work}\label{sec:relwork}
The use of market-based approaches for the decentralized control of \ac{DER} is not a novel practice. 
An early work about scheduling household appliances using computational markets was presented by Ygge in \cite{YggeA96}.
A survey of demand response solutions is presented in \cite{palensky}, while challenges towards the development of market platforms and intelligent agents for smart grid control are discussed in \cite{Ramchurn:2012:PSS,powermatcher,Lamparter}.
So far different mechanisms have been used to trade energy \cite{saad:magazine},
cooperative games~\cite{Alam:2013},
as well as based on cost-minimization and non-cooperative games~\cite{powermatcher,adika,chavali,5628271} especially
double auctions (DA) \cite{Vytelingum:2010,nobel,doubleauctiongametheory}.
While energy trading in power grids has been widely used as a way of coordinating energy resources beforehand, a methodology to trade and therefore manage energy in smaller systems, such as microgrids and smart buildings, is yet to be investigated.
In particular, there has been little research in designing systems considering users' behavior and preferences.
%
In \cite{Bapat:2011}, a system is designed to determine preferred time of use of appliances to minimize running costs and activity disruptions.
More recent works, such as iDR~\cite{Chandan:2014}, DRSim~\cite{drsim} and the HEMS simulator~\cite{monacchi2014:hems},
aim at extracting users' consumption behavior and preferences in order to minimize the discomfort produced by control strategies.
This work differentiates from previous studies, by addressing the stringent requirements and high volatility in power supply and demand typical of microgrids.

\section{Problem statement}\label{sec:problem}
This study concerns the design of a microgrid controller, that is, an agent controlling the energy being exchanged with and throughout the local power system.

\subsection{Modeling electrical loads}
An electrical load is described by a price sensitivity model $\psi$ denoting the maximum price per $kWh$ users are willing to pay to operate the device.
Each appliance is a collection of services, each described by an \textit{operation model} and a \textit{usage model} \cite{monacchi2014:hems, egarter2014integration}.
The operation model describes the coordinated execution of the system components in terms of a state sequence, in which a state $\sigma_{i}$ is defined as a peak power level $P_{i} \in \mathbb N^+$ and a duration $d_{i} \in \mathbb N^+$ in seconds.
In addition, each state is associated to a \textit{start delay sensitivity} modeling the maximum tolerated start delay in seconds and an \textit{interruption sensitivity} $\chi^i$ defining the severity under which the state operation can be interrupted.
The start delay sensitivity models the \textit{device start delay sensitivity} $\chi^s$ for the initial state, and the \textit{state start delay sensitivity} $\chi^s$ otherwise.
Those values respectively affect user's comfort and the correct device operation.
It is also important to remark that appliance services, based on their start delay sensitivity, are classified as flexible or inflexible.
Delayed start results in a much higher utility loss for inflexible devices, as resulting from an unconstrained price sensitivity.
The usage model defines the time of use proability, namely the probability to start the device at a specific time of the day.
Such willingness value $\omega^* \in [0,1]$ is associated to a willingness decay $\lambda \in \mathbb R$, which updates the probability based on concluded device operations.
As shown in \cite{monacchi2014:hems}, those values can be either statically defined or directly extracted from a consumption dataset using appropriate tools such as \cite{umma}.
In \cite{greend}, a Bayesian network was used to model appliance usage behavior in an Austrian household.
Appliances which miss a network interface or are not subject to flexible control are subsumed as inflexible devices and will be charged under the grid price.
This offers an incentive for coordination and allows for handling a mixture of smart and non-smart device.
In order to achieve a more efficient energy management, \cite{egarter2014integration} show how to detect non-smart (legacy) devices and automatically fill a machine-readable profile.

\subsection{Designing a power broker}
Energy brokerage is the problem of formulating a price describing the cost associated to the provisioning of the currently available power.
We consider the power grid and the local generators as truth-telling agents,
whose reservation price is given by the grid-energy tariff and the feed-in tariff, hereby represented as \textit{get} and \textit{fit}.
For the broker, the cost $p$ to supply the local grid at a given instant with a required power $P_s$ is thus given by the power drawn from the local generator, $P_{re}$, and the power requested from the energy grid, as in Eq.~\ref{eq:price}.
\begin{equation}\label{eq:price}
p(P_{s}) = \left\{
  \begin{array}{l l}
    fit & \quad \text{if $P_{s} \leq P_{re}$}\\
    \frac{P_{re} \cdot fit + (P_{s}-P_{re}) \cdot get}{P_{s}} & \quad \text{if $P_{s} > P_{re}$}
  \end{array} \right.
\end{equation}
where $P_s = P_{re} + P_{grid}$.
%
\begin{figure}[h!]
	\centering
	\includegraphics[width=0.8\columnwidth]{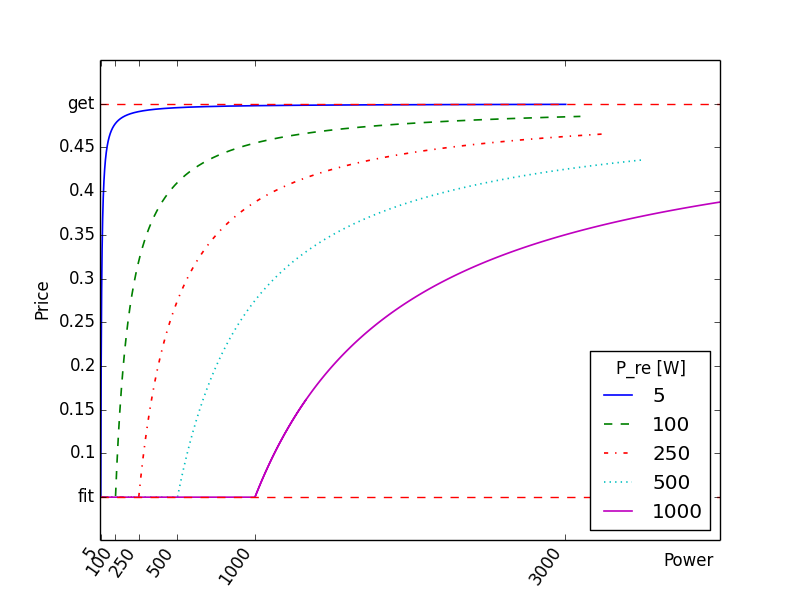}
	\caption{Cost function for different amounts of locally generated power $P_{re}$}\label{fig:price_model}
\end{figure}
Fig.~\ref{fig:price_model} shows a scenario with $3000W$ provided by the power grid, under a $0.05~€/kWh$ feed-in tariff and a $0.5~€/kWh$ grid energy tariff.
The price is computed for different levels of $P_{re}$.
Clearly, a higher amount of renewable power lowers the portion of grid power being used, with a consequent lower price to supply the local grid.
However, as previously discussed the broker is required to also minimize service interruptions by providing multiple provisioning durations or service-level agreements (SLA).
Trading different durations as different products can better reflect demand differences for the service agreements.
For instance, with a majority of loads with long states this would imply higher costs to purchase long-term service agreements, which would favour the purchase of short-term contracts and the mitigation of fragmentation.
\begin{figure}[h!]
	\centering
	\includegraphics[trim=4.65cm 10.4cm 12.75cm 15.5cm,clip,width=0.65\columnwidth]{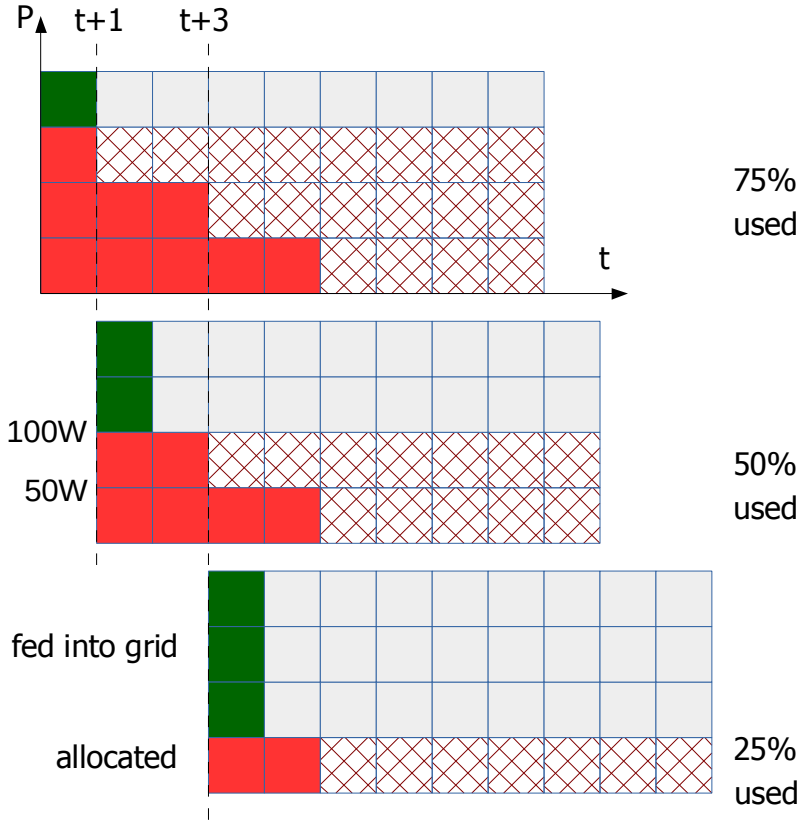}
	\caption{Providing multiple provisioning contracts}
	\label{fig:slas}
\end{figure}
The example shows three loads of same type, and consequently same demand (i.e., $50 W$), competing for the allocation of power at time $t$.
At time $t+1$ the running loads cause a $150 W$ demand, while the remaining power is fed into the grid.
As time passes and no new allocations are matched, the service-level agreements are shifted to the left, thus resulting in the situation showed in the second and third example.
The broker's objective is to maximize the profit, indicated as difference between its income and costs (i.e., to buy energy from the grid and the local generator).
This includes a profit $\Pi_{uGrid}$ resulting from power sold throughout the microgrid, as well as $\Pi_{feedin}$ resulting from power injected back to the main power grid.
Similarly, we distinguish in a procurement cost $C_{supply}$ and a compensation cost $C_{reimbursement}$.
Agreements that can not be satisfied due to insufficient supply are reimbursed.
Specifically, the broker refunds involved loads with the supply cost for the remaining portion of the SLA.
The overall broker profit $\Pi$ is thus given as: \begin{equation}\Pi = \label{eq:fitness} (\Pi_{uGrid} + \Pi_{feedin}) - (C_{supply} + C_{reimbursement}) \end{equation}
\subsection{Efficient Power allocation}
\begin{figure}[h!]
	\centering
	\includegraphics[trim=5.818cm 22.1cm 9.831cm 6.45cm,clip,width=0.6\columnwidth]{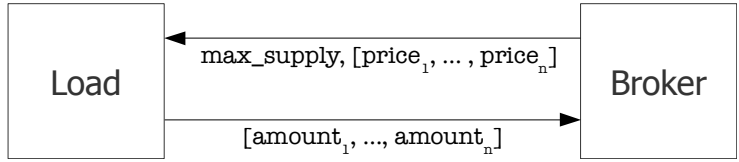}
	\caption{The allocation mechanism}\label{fig:allocationmechanism}
\end{figure}

To identify the load that best suits the formulated price vector, i.e., the one that associates highest value to the available power, we can employ market-based mechanisms.
One possibility is to consider the formulated price vector as reservation price and use a price-ascending auction to select the load that values the resource most.
Price-ascending auctions work on multiple rounds where the price is increased as long as the demand is more than available supply.
Bidding over multiple iterations sharing a tentative unit price allows for a progressive reduction of price uncertainty, a property called price discovery.
Given that a customer rationally drops out the auction at a certain price, we can estimate its utility to better act in future markets.
This consequently results in a lighter cognitive process to decide the quantity to purchase \cite{Ausubel04auctioningmany}.
%
To focus on the broker design, we do not employ any market-based allocation at the current stage.
We directly allocate a SLA for those customers willing to pay the price formulated by the broker.
This removes the burden, in terms of both communication and time, to run an auction over multiple consecutive rounds.
In addition, this eases the design of bidders.
Whilst strategic bidding is necessary to achieve user-centric control and effectively meet comfort levels, we focus hereby on system-wide performance achievable through the broker.
Each load is thus a reactive agent which is given a price vector and replies with a quantity vector (see Fig.~\ref{fig:allocationmechanism}).
The quantity vector models the demand of the load for those SLAs whose duration suits the one of the operational state.
Specifically, the load selects a pool of candidate SLAs among those respecting its utility (i.e., sensitivity price), sorts them by price and selects the best candidate according to a policy.
In particular, we employ a best-fit strategy in which the load aims at allocating the shortest SLA which is big enough to contain the state.
%
%
\section{Experiment}\label{sec:experiments}
\subsection{Evaluation metrics}
In this study we seek to compare the performance of different brokerage schemes.
In particular, we have identified the following metrics:
\begin{description}
  \item[\textbf{Peak-to-average ratio}] The PAR is computed as ratio of the peak power to the average over the considered time window.
  This factor describes the proportion of power peaks over the overall demand and directly affects the loss of load probability.
  It is thus desirable to keep this value as low as possible, as a high PAR determines a lower system reliability and consequently inefficiency \cite{6840951}.
  \item[\textbf{Service availability}] The availability is the proportion of time in which a system is in a working condition.
  For clarity we distinguish in two more measures: the \ac{MTBF} which models the average uptime between consecutive failures, and the \ac{MTTR} describing the average downtime due to a service recovery.
  The \ac{MTBF} is directly related to the failure rate (i.e, the frequency of service interruption) as $\lambda = \frac{1}{MTBF}$.
  The availability is computed as $A = \frac{MTBF}{MTBF+MTTR}$, while the unavailability is $U = 1 - A =\frac{MTTR}{MTBF+MTTR}$.
  \item[\textbf{System reactivity}] The reactivity of a system describes its degree of responsiveness,
  which in the context of energy management can be defined for a load as the probability of having enough power to operate.
  We can thus distinguish in: i) $CBP$ number of times the load could make an offer to get enough power and ii) $CNBP$ number of times the load could not make an offer to get enough power.
  These values can be collected for each load once for each trading day.
  Consequently, we can compute $R = \frac{CBP}{CBP+CNBP}$.
  Clearly, the presence of longer service agreements increases the proportion of $CNBP$ with respect to $CBP$, thus lowering $R$.
  \item[\textbf{Economic profit} ($\Pi$)] The profit in economical terms is computed as difference between retail revenues and production costs.
  Profit maximization is the driving force of capitalistic markets.
  This is directly proportional to a firm's market power, the ability to raise the price of a good or service over its marginal cost.
  Clearly, this is high in monopolies and oligopolies, and absent in perfectly competitive markets.
  In this study, the broker operates in a pure monopoly and determines the price of the commodity to steer the system.
  Consequently, the economic profit represents a good quality measure of the broker's performance.
\end{description}
\subsection{Research questions}
As previously introduced, the broker seeks profit maximization by correctly formulating the retail price of multiple provisioning durations.
For costs are fixed by the \textit{get} and \textit{fit} plans, this translates into the correct reflection of future supply costs and expected demand into the retail price.
This directly leads to distinguish in: i) a \textit{pessimistic} broker that charges SLAs proportionally to their duration (the longer the reservaton, the higher the price),
and ii) an \textit{optimistic} broker that keeps the same price for all SLA durations.
These strategies are useful as baseline reference.
Accordingly, the pessimistic broker matches mostly single-unit SLAs (i.e., the shortest available).
This highlights the problems of market competition and service interruption that motivate this study.
On the opposite side, the optimistic broker attributes same uncertainty to all SLA durations.
Intuitively, such a naive price formulation might lead to economic losses for the broker.
Moreover, this can result in the a high number of long-term SLAs, which reduces market competition and consequently affects the reactivity of the allocation mechanism.
%
%
This study assesses multiple brokerage approaches for microgrids, by addressing the following questions:
\begin{enumerate}
  \item \textbf{Does the use of service-level agreements actually mitigate service interruption?}
  A pool of provisioning durations shall be selected to model service-level agreements for a microgrid.
  The effect of using SLAs over the rate of service interruption is analyzed and discussed.
  \item \textbf{How does the system reactivity change given the availability of service-level agreements?}
  Provisioning agreements engage the system by reserving future available supply.
  The cost of this approach in terms of system reactivity should be further analyzed.
  \item \textbf{How does a model of supply and demand improve power pricing?}
  As already shortly introduced, a model of the environment could be exploited by the broker to better price available power.
  In fact, the model can be learned from previous interactions within the microgrid.
  Possible representations and their effectiveness in terms of market profit should thus be investigated.
\end{enumerate}
\subsection{Setup}
\begin{figure}[h!]
\centering
	\begin{subfigure}[b]{0.45\columnwidth}
		\centering
		\includegraphics[width=\columnwidth]{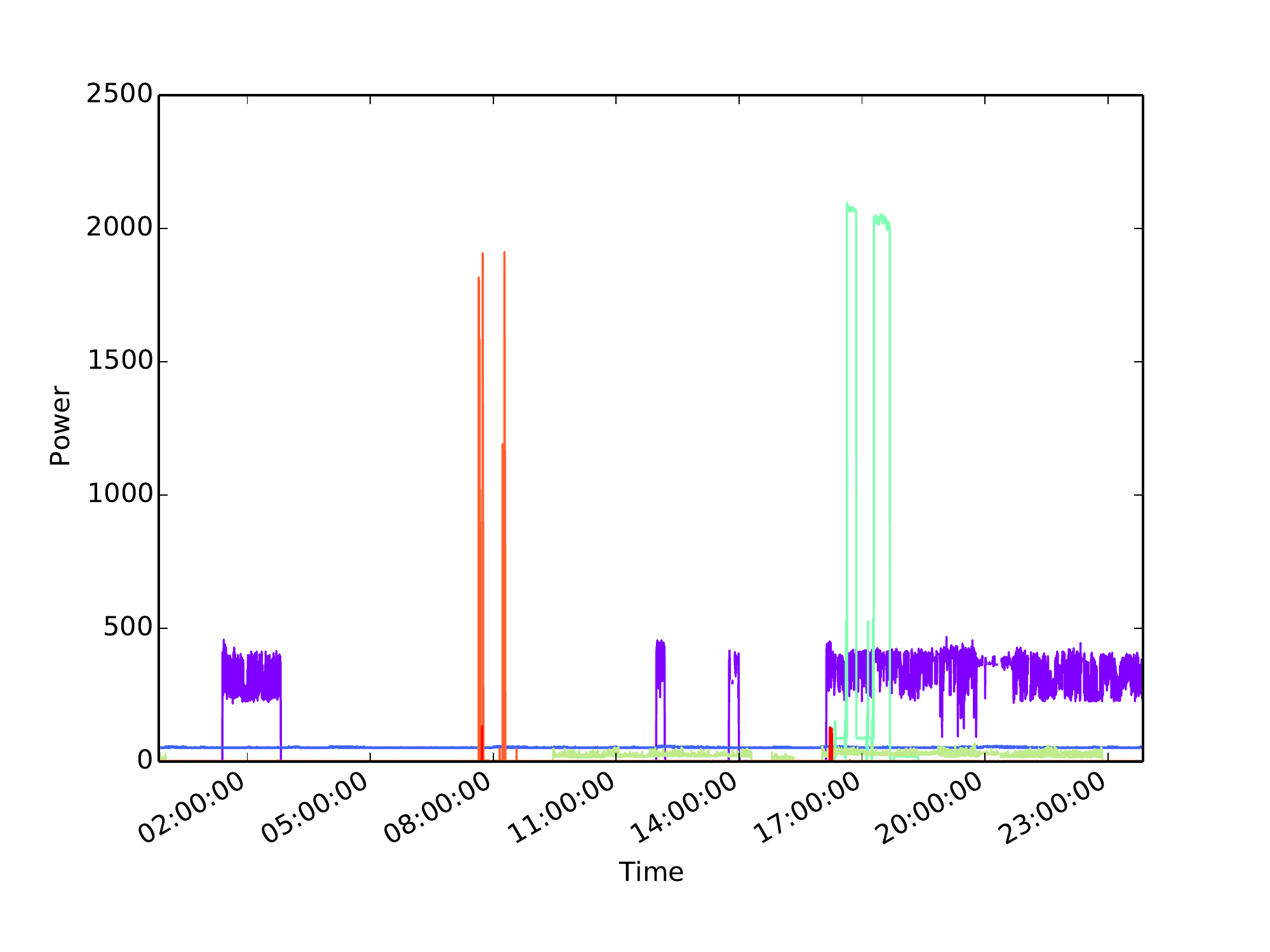}
		\caption{Data as provided by the GREEND}
	\end{subfigure}
	~
	\begin{subfigure}[b]{0.45\columnwidth}
		\centering	
		\includegraphics[width=\columnwidth]{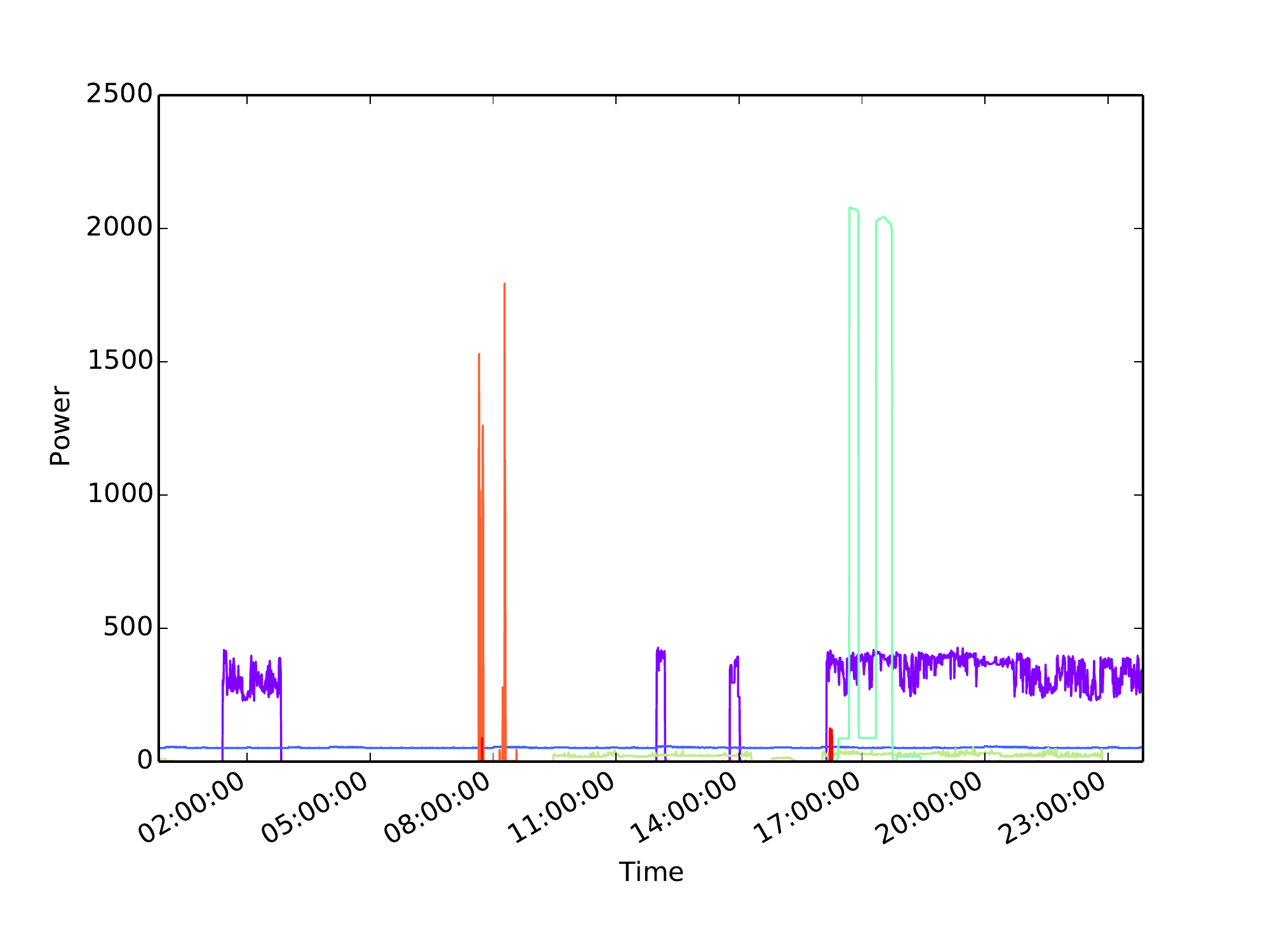}
		\caption{Data after the preprocessing stage}
	\end{subfigure}
	~
	\begin{subfigure}[b]{0.45\columnwidth}
		\centering	
		\includegraphics[width=\columnwidth]{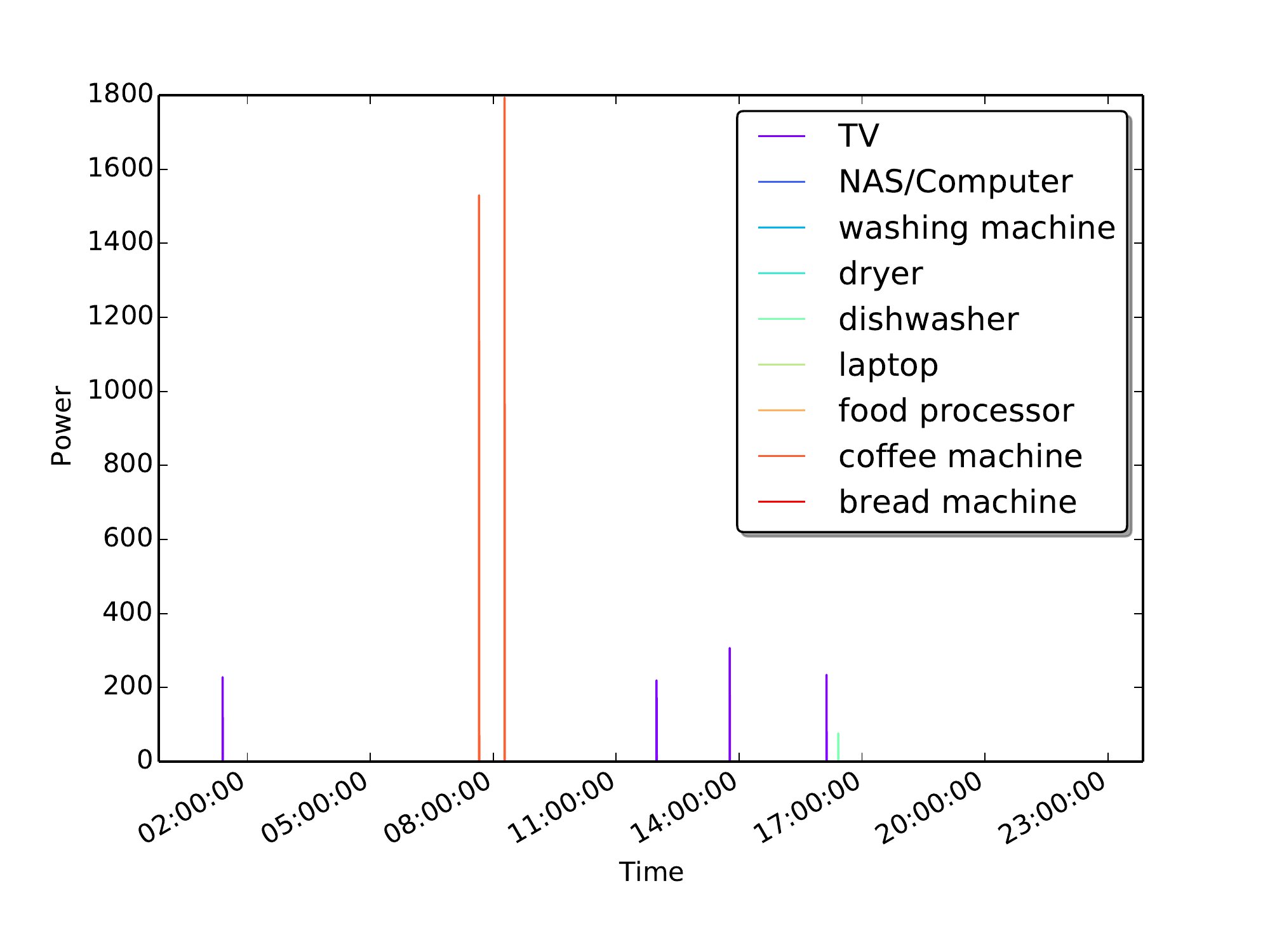}
		\caption{Detected events}\label{fig:events_detected}
	\end{subfigure}
	\caption{Usage behavior of January 1st 2015 for the selected household}\label{fig:usage_model_benjamin}
\end{figure}

The performance of our approach depend on the consumption scenario involved, that is, the number and type of loads, their usage model, and so on.
Our scenario models a small Austrian household with a pool of appliances and a photovoltaic power generator \cite{monacchi:2013Nov}.
The household is connected to a 5~kWp photovoltaic plant.
To fully control the simulation and assess different scenarios, we use synthetic power data computed through models from the RAPSim~\cite{manfred2014}.
In detail, the calculation considers latitude, longitude, area and efficiency of the equipment, as well as a cloud coverage model to compute the sunlight intensity.
Consumption data are derived from building 2 of \cite{greend} which models the usage behavior of a retired couple with an adult son.
In particular we selected the following devices: television, dishwasher, tumble dryer, washing machine, fridge and coffee machine.
A low-pass filter was applied to the measurements and edge-detection techniques were employed to identify device starting events (See Fig.~\ref{fig:usage_model_benjamin}).
Each load operation was ultimately described as a sequence of states (Table~\ref{tab:scenario}) accompanied by an external usage timeserie.
For simplicity, the fridge was modeled as a periodic device with $\omega^* = 0.8$ and $\lambda = 0.5$.
\begin{table}[h!t]
 \centering
 \caption{simulation scenario}\label{tab:scenario}
	\begin{tabular}{| l |}
    \hline
    	\textbf{Operation model} $(P [kW], d [sec])$	\\
    \hline
   		$(0.18, 3600)$\\
   		$(2.1,300), (0.1,120), (0.3,60), (0.1,120), (2.1,300)$\\
   		$(2.5, 120)^{10}$\\
   		$(2.1,120), (0.3,300), (0.2,120), (0.6,300), (0.2,60)$\\
   		$(0.2,30), (0.16,600)$\\
   		$(2, 60)$\\
    \hline
    \end{tabular}
\end{table}
All models were then implemented in the HEMS simulation package \cite{monacchi2014:hems}, an extension of the FREVO framework for evolutionary design~\cite{istvanfrevo}.
This allows for a quick implementation of the scenario and the later employment of an artificial neural network for our forecasting purposes.
%
%
\section{The value of coordination and provisioning contracts}\label{sec:coordination}
\begin{figure}[h!]
	\centering	
	\begin{subfigure}[b]{0.75\columnwidth}
		\includegraphics[trim=0cm 0.2cm 0cm 0cm,clip, width=\columnwidth]{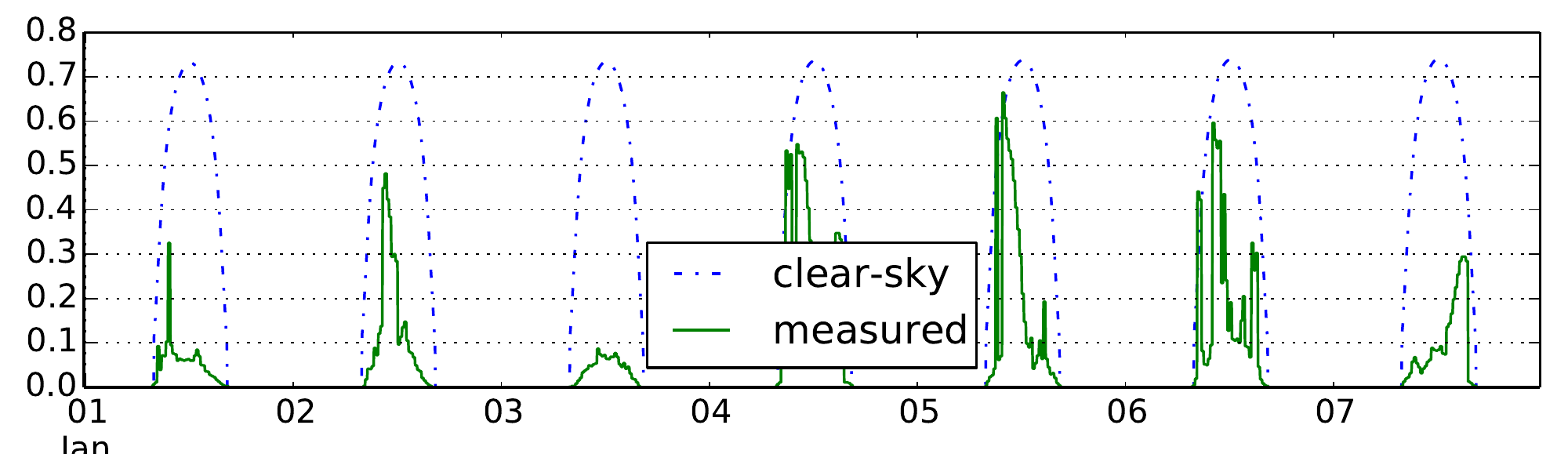}
		\caption{January 2015}
	\end{subfigure}
	\begin{subfigure}[b]{0.75\columnwidth}
		\includegraphics[trim=0cm 0.2cm 0cm 0cm,clip, width=\columnwidth]{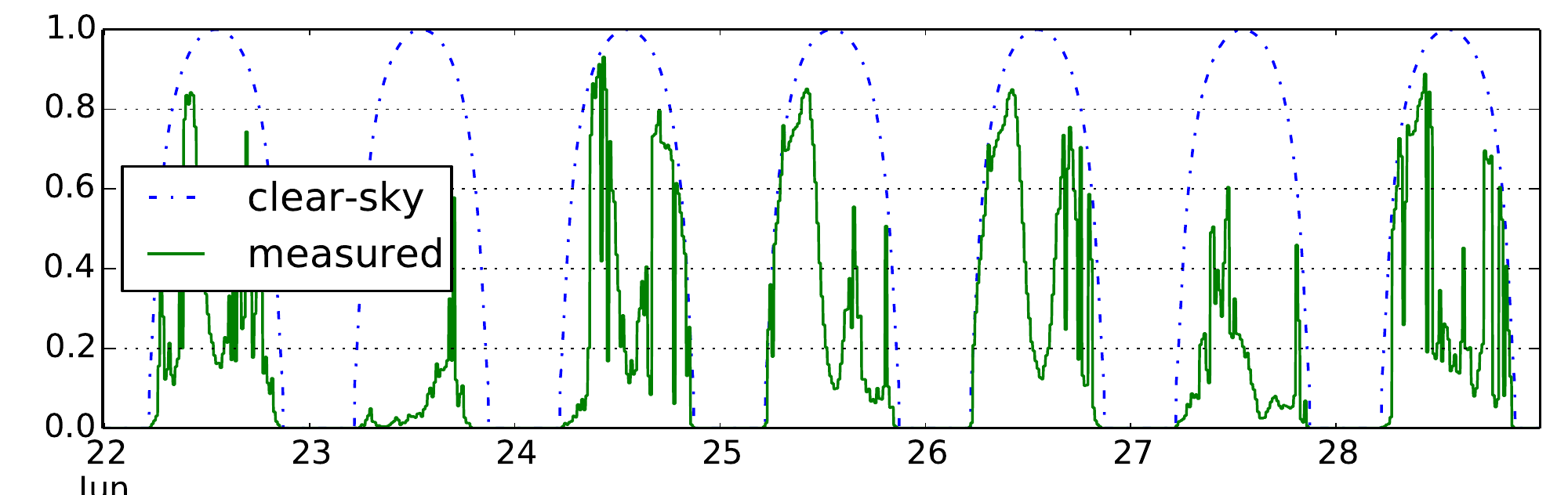}
		\caption{June 2015}
	\end{subfigure}
	\caption{Sunlight intensity from the employed models}\label{fig:weather}
\end{figure}

To evaluate the optimistic and pessimistic brokers we firstly disable the market-based allocation by setting the price sensitivity of all loads to 0.9~€/kWh.
Selected provisioning durations are: a unitary agreement occupying 1 time instant, and respectively 10, 30, 60, 120, 600 and 1800 seconds.
The scenario was simulated according to the power measurements of building 2 in the first week of 2015 (i.e., Jan 1st to 7th).
We then varied the amount of $P_{grid}$ available to the broker: Plan1: 0~kW, Plan2: 1.5~kW, Plan3: 3~kW in 6 a.m. to 6 p.m. and 1~kW otherwise, Plan4: 3~kW and Plan5: 6~kW.
The photovoltaic plant was sized to 3.3~kWp.
The production depends on two different weather models (See Fig.~\ref{fig:weather}): i) a clear-sky sunlight intensity based solely on the sun position,
and ii) a 15-minute-resolution illuminance timeseries collected from a weather station at the University of Klagenfurt\footnote{http://wetter-cms.aau.at/info.php}, Austria.
The current Austrian energy pricing system does not implement time-based tariffs to reflect energy shortage, mainly for the absence of digital meters. 
To model a future scenario, we assumed time-based tariffs similar to the Italian tariffs with $get$ being 0.29 €/kWh in the interval 6 a.m. to 9 p.m and 0.15 €/kWh otherwise.
The feed-in tariff $fit$ is 0.04 €/kWh from 6 a.m. to 9 p.m. and 0.02 €/kWh otherwise.
\begin{figure}
	\centering
	\begin{subfigure}[b]{0.45\columnwidth}
		\includegraphics[trim=1.8cm 1cm 2.035cm 1.5cm,clip, width=\columnwidth]{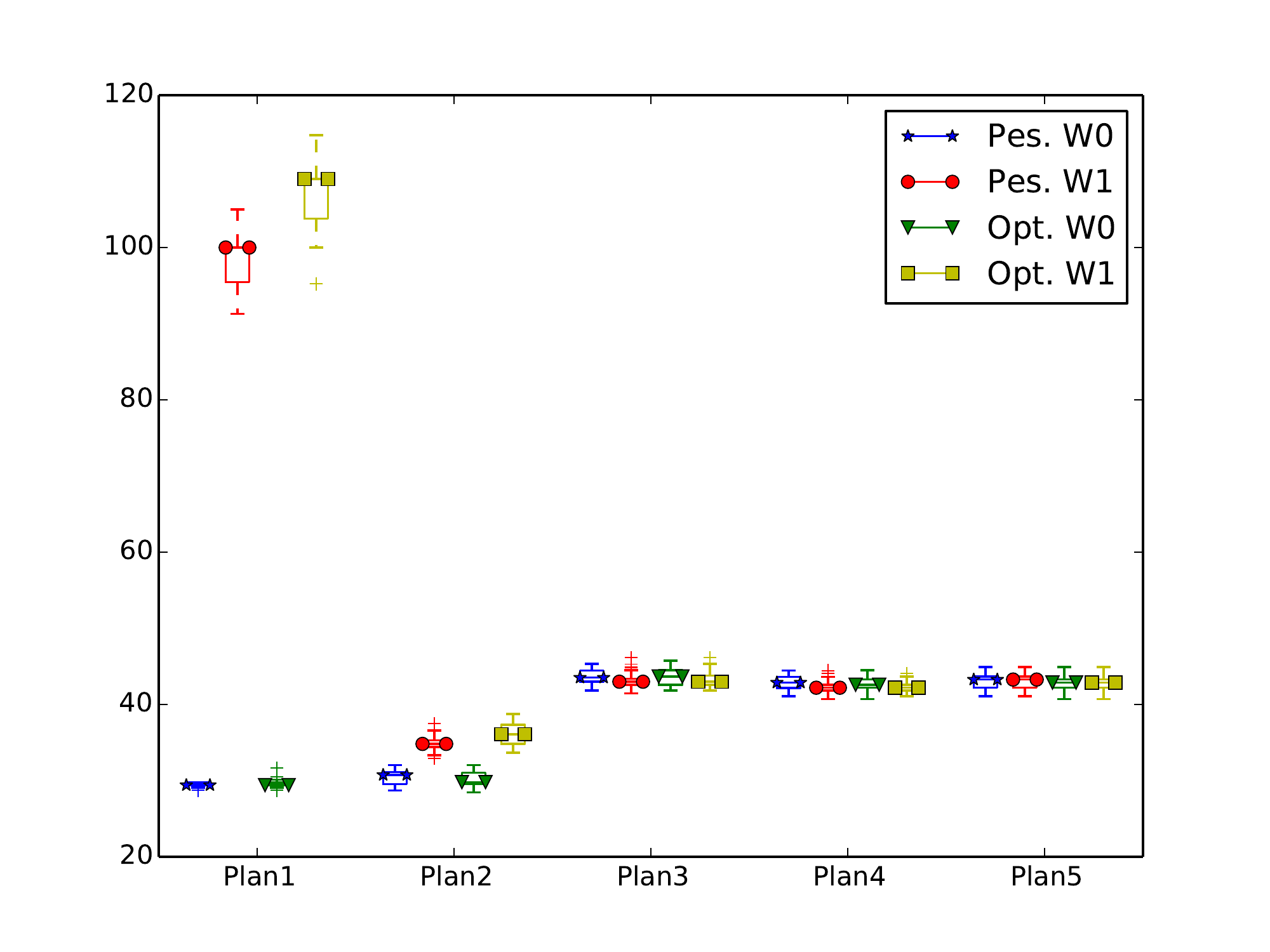}
		\caption{PAR for different $P_{grid}$ scenarios}
	\end{subfigure}
	~
	\begin{subfigure}[b]{0.45\columnwidth}
		\includegraphics[trim=1.8cm 1cm 2.035cm 1.32cm,clip, width=\columnwidth]{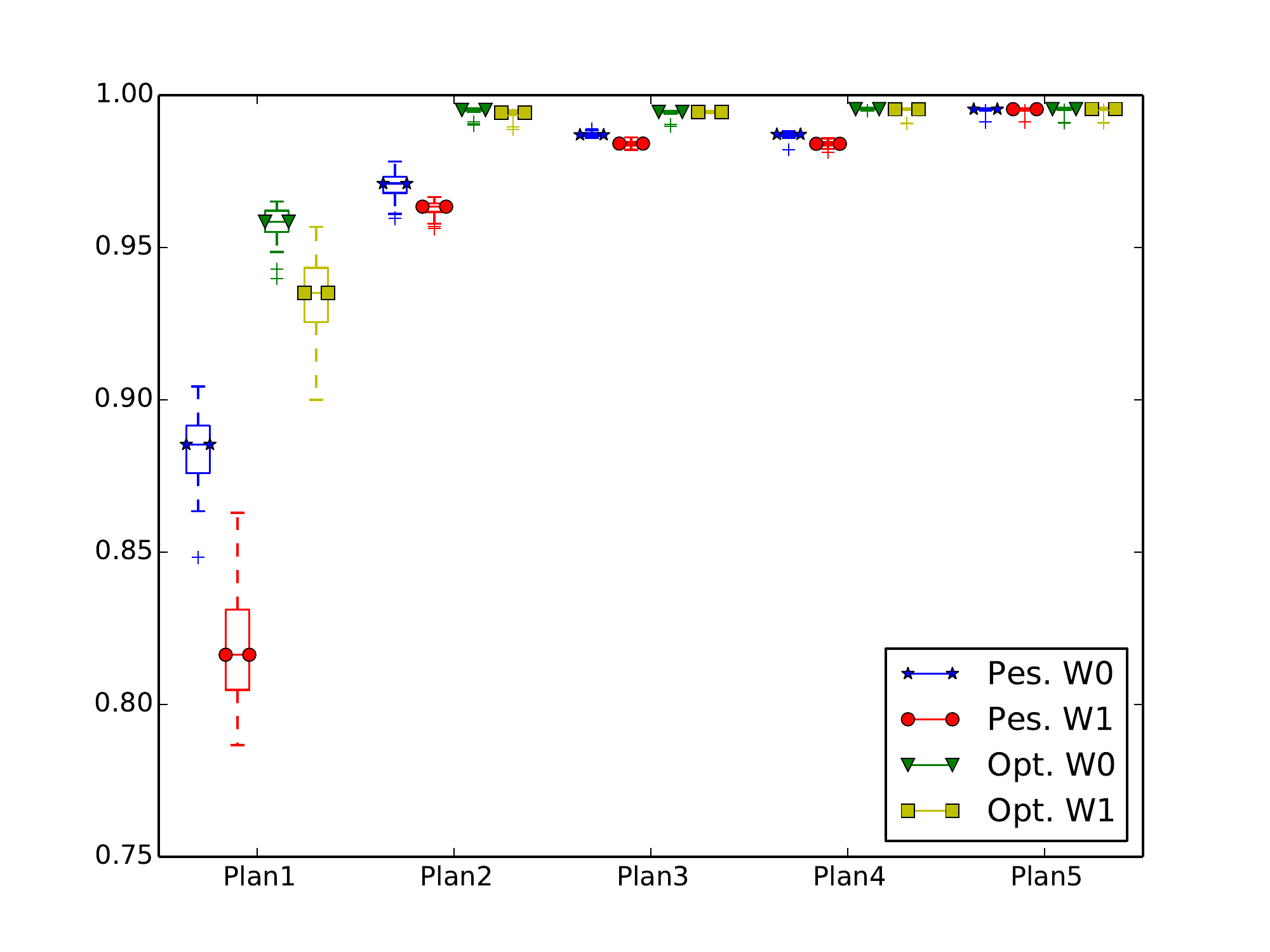}
		\caption{Availability for different $P_{grid}$}
	\end{subfigure}
	
	\begin{subfigure}[b]{0.45\columnwidth}
		\includegraphics[trim=1.8cm 1cm 2.035cm 1.32cm,clip, width=\columnwidth]{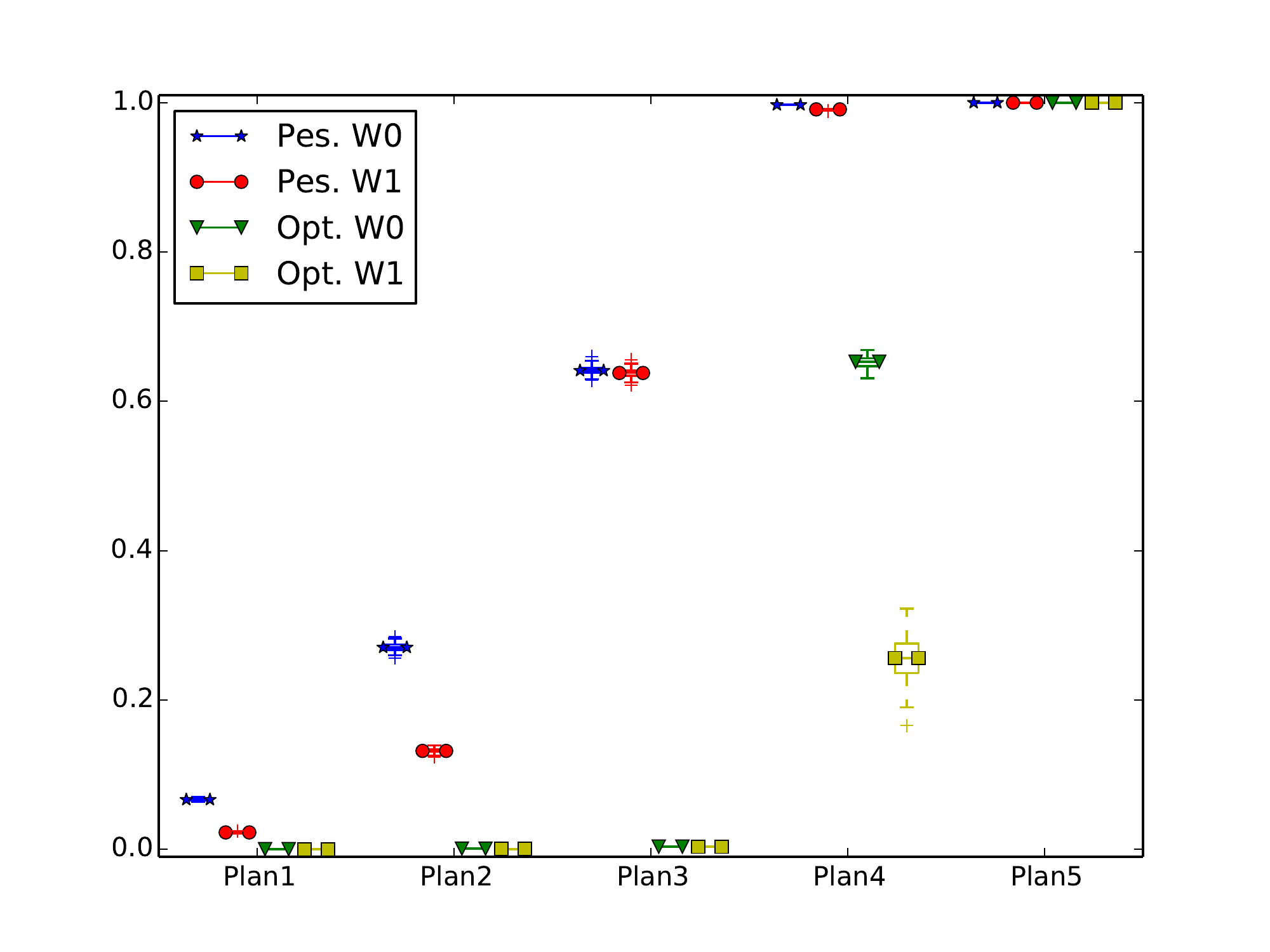}
		\caption{Reactivity for different $P_{grid}$ scenarios}
	\end{subfigure}
	~
	\begin{subfigure}[b]{0.45\columnwidth}
		\includegraphics[trim=1.8cm 1cm 2.035cm 1.5cm,clip, width=\columnwidth]{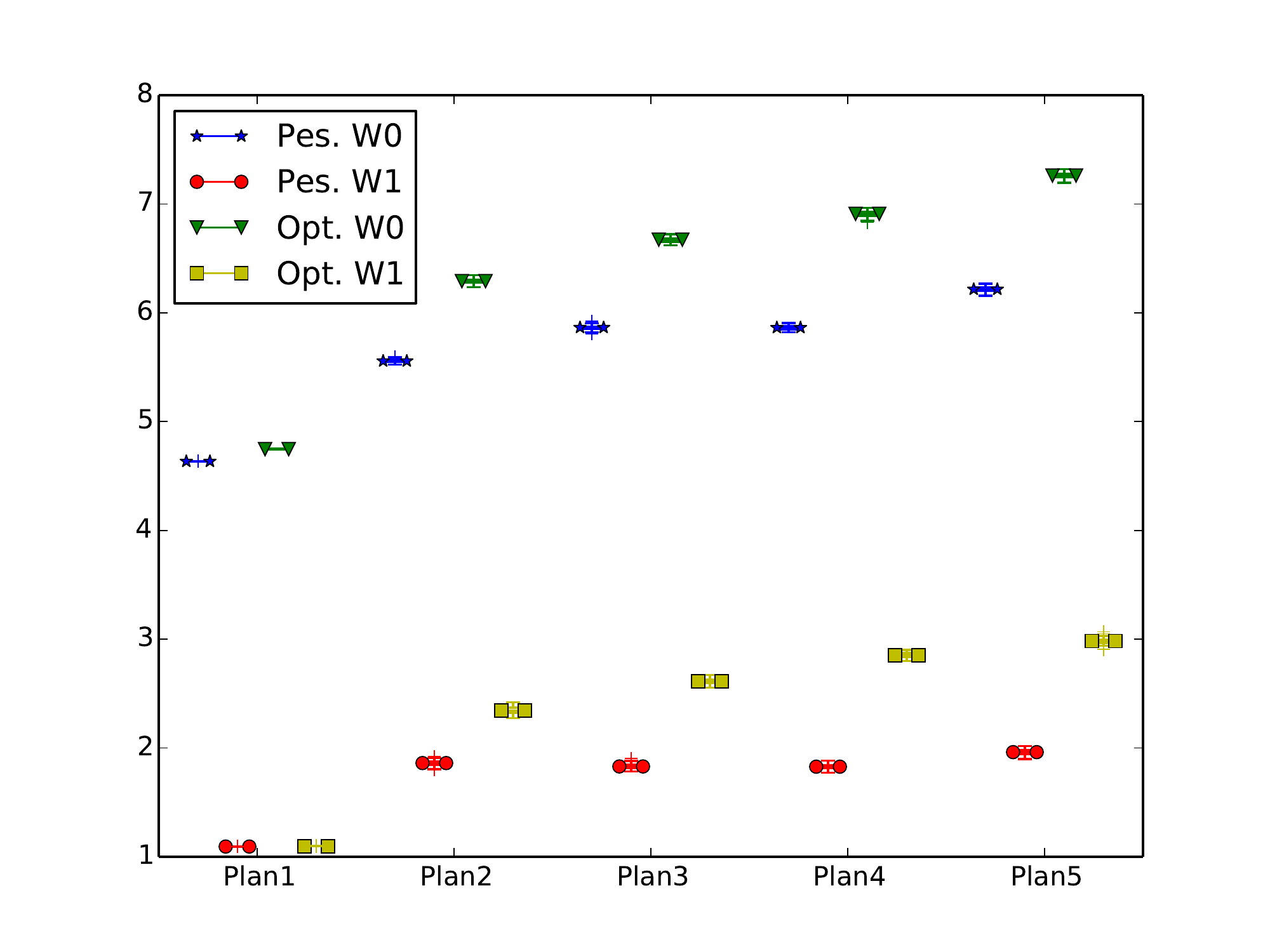}
		\caption{Broker's profit for different $P_{grid}$ scenarios}
	\end{subfigure}
	\caption{Evaluation metrics for 100 evaluations}\label{fig:evaluation_rule_based}
\end{figure}
\begin{figure}
	\centering
	\begin{subfigure}[b]{0.45\columnwidth}
		\includegraphics[trim=1.8cm 1cm 2.035cm 1.32cm,clip, width=\columnwidth]{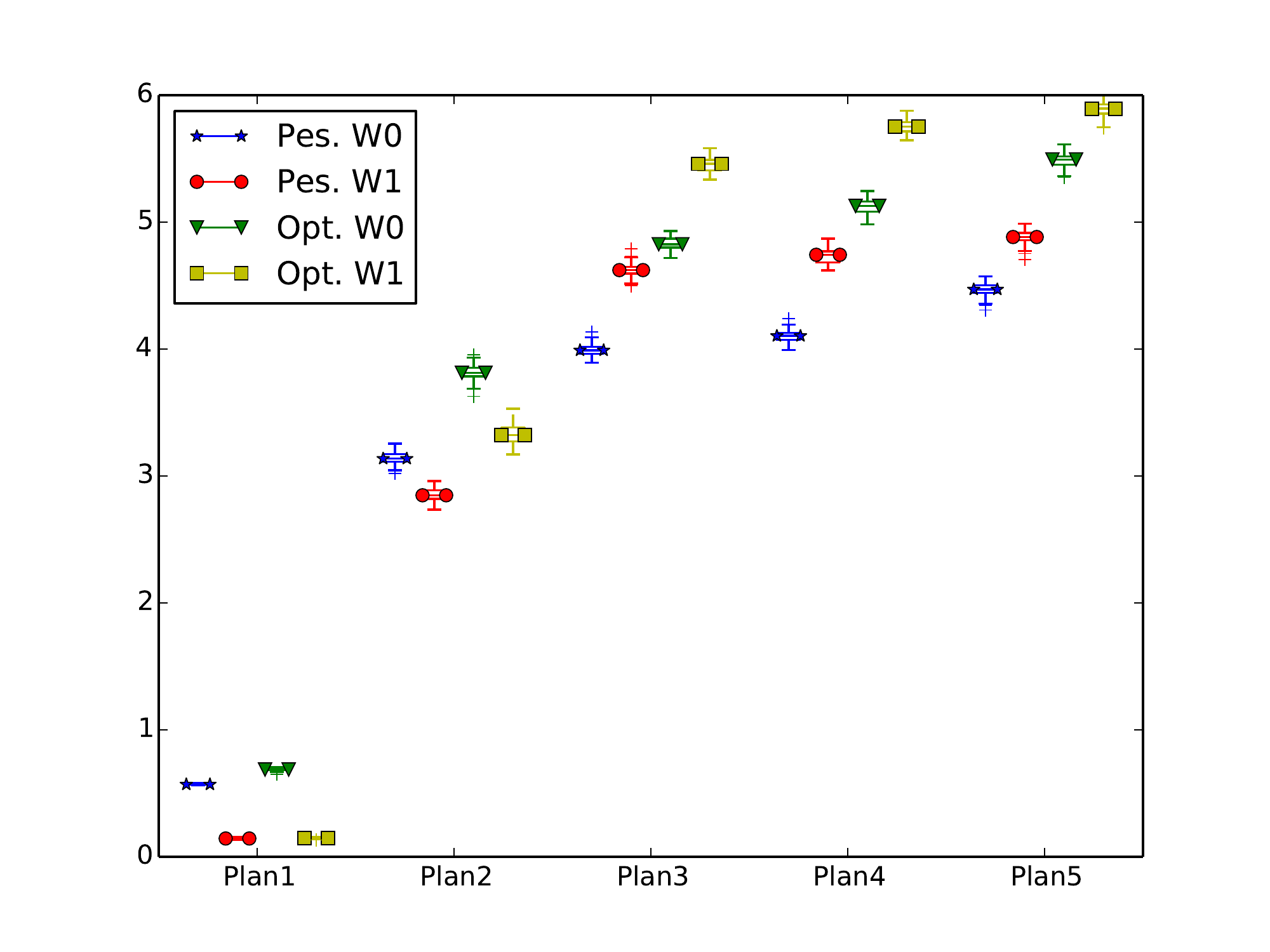}
		\caption{Income from local sale for different $P_{grid}$ }
	\end{subfigure}
	~
	\begin{subfigure}[b]{0.45\columnwidth}
		\includegraphics[trim=1.8cm 1cm 2.035cm 1.32cm,clip, width=\columnwidth]{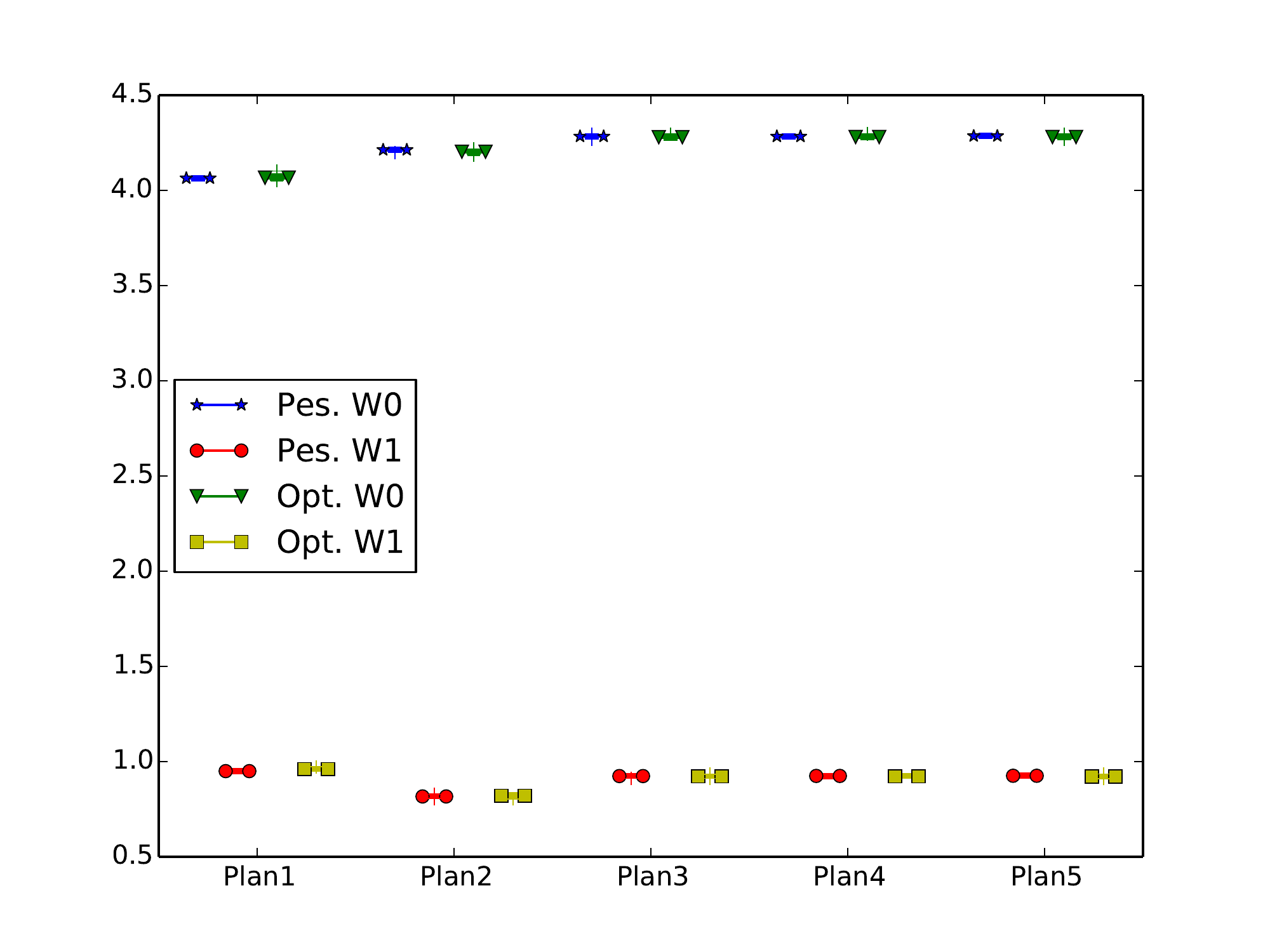}
		\caption{Income from feed-in for different $P_{grid}$ }
	\end{subfigure}
	
	\begin{subfigure}[b]{0.45\columnwidth}
		\includegraphics[trim=1.8cm 1cm 2.035cm 1.5cm,clip, width=\columnwidth]{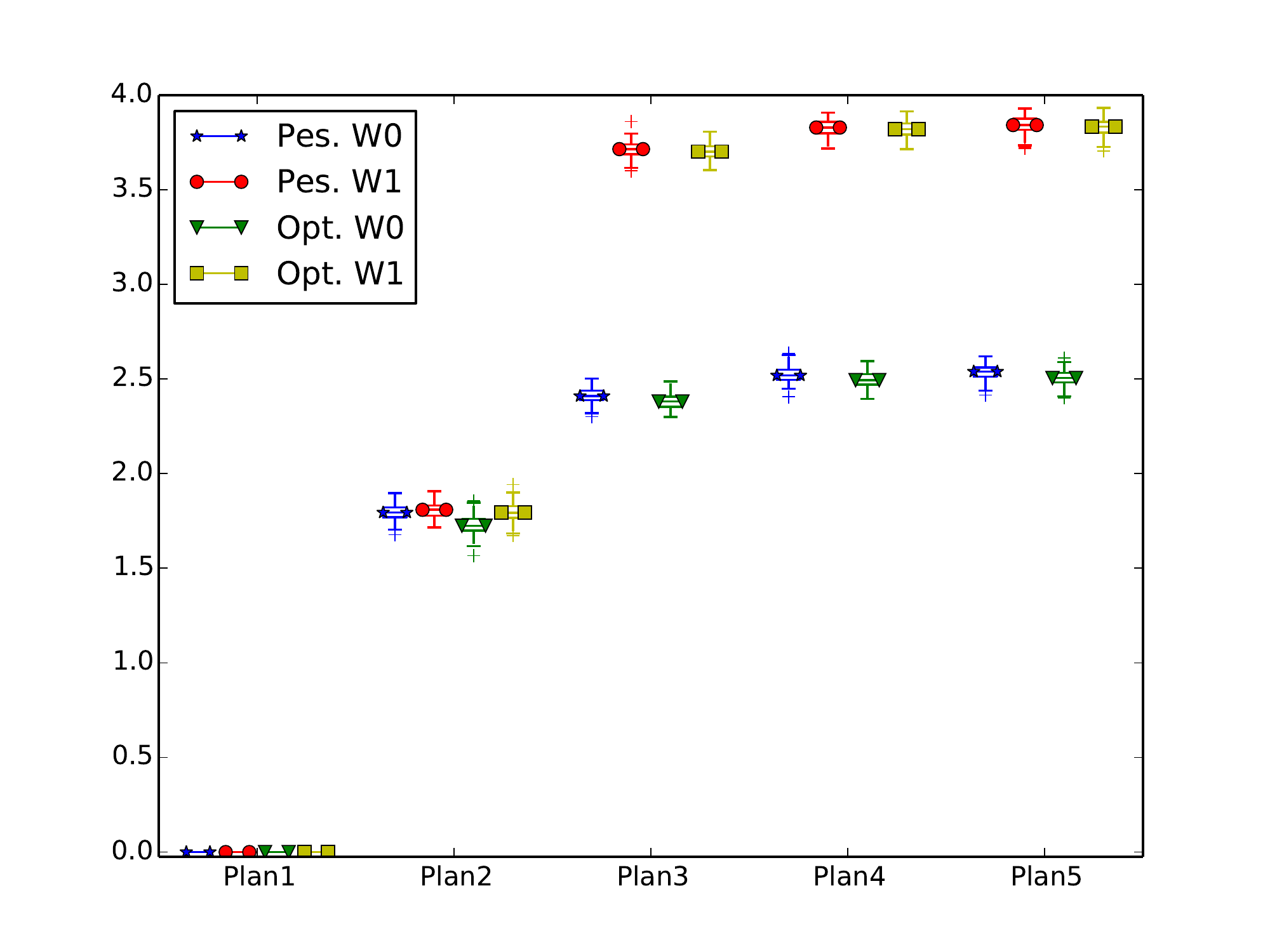}
		\caption{Supply costs for different $P_{grid}$ scenarios}
	\end{subfigure}
	~
	\begin{subfigure}[b]{0.45\columnwidth}
		\includegraphics[trim=1.8cm 1cm 2.035cm 1.35cm,clip, width=\columnwidth]{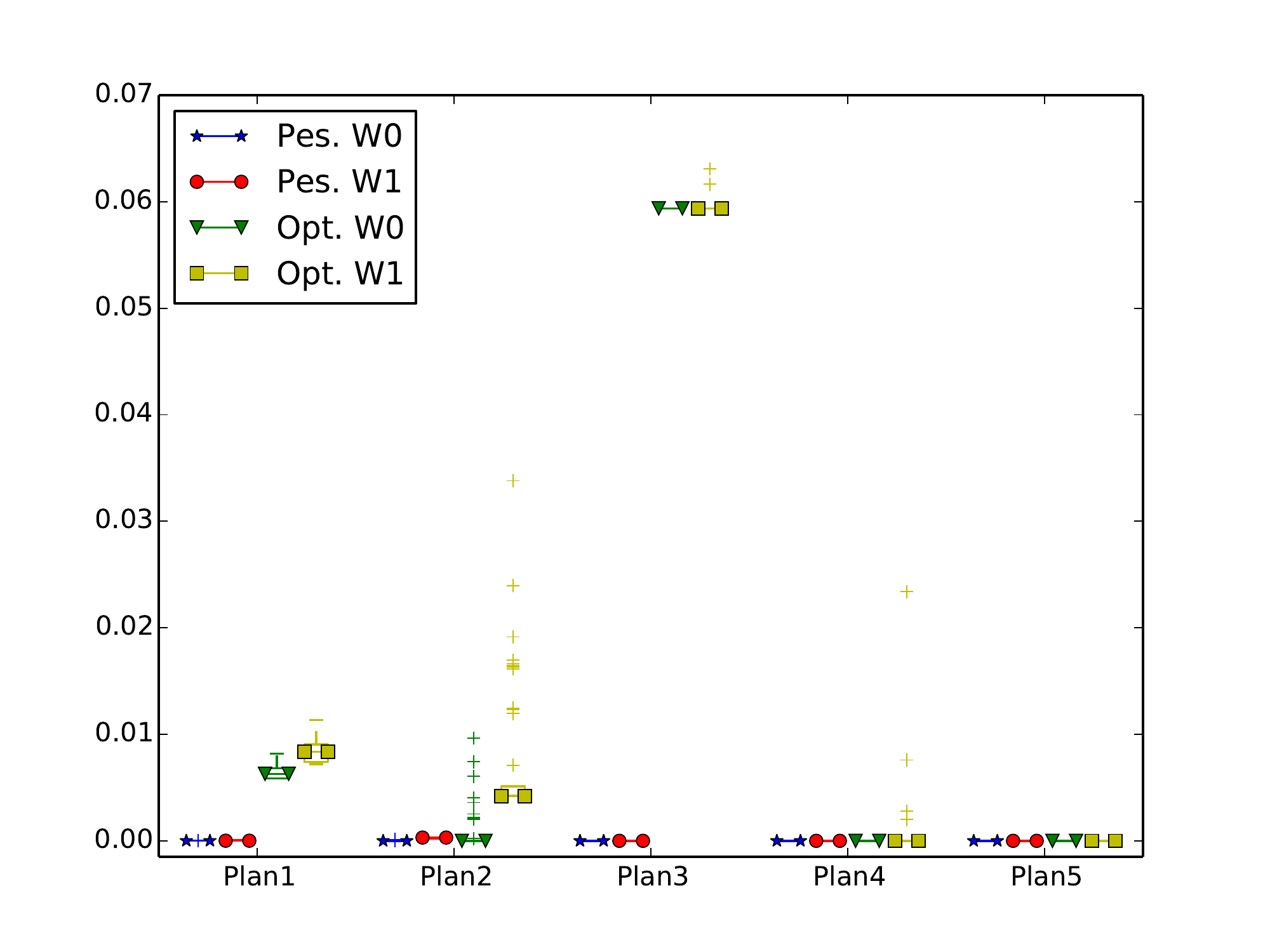}
		\caption{Reimbursement costs for different $P_{grid}$ }
	\end{subfigure}
	\caption{Profit components for 100 evaluations}\label{fig:evaluation_rule_based_2}
\end{figure}
As visible in Fig.~\ref{fig:evaluation_rule_based},
lowering $P_{grid}$ causes the postponement of loads to off peak periods, but also prevents some loads to run at all. While the former reduces the peak power demand, the latter also decreases the average power demand, thus leading to the PAR reported.
The postponement of devices is not directly reflected on the availability $A$, as this captures only the performance of operating loads.
For instance, the fridge and the entertainment system are the only operating loads in the first scenario.
Hence, lowering $P_{grid}$ has the effect of preventing the operation of certain high-power demanding loads, which results in a lower profit $\Pi_{uGrid}$.
%
Moreover,
it is remarkable that availability and reliability are two opposite objectives to be optimized.
Specifically, with a low $P_{grid}$ the sale of long-term provisioning agreements results in resource monopolization (i.e., $R \simeq 0$).
The absence of a connection to the main power grid, as in the first scenario, makes the system more sensitive to variations of $P_{re}$.
The broker reimburses involved loads with the supply cost of the remaining portion of the SLA.
The very low values are due to the pricing of the SLA, which in absence of $P_{grid}$ is charged under the $fit$ tariff (See Eq.~\ref{eq:price}).
As visible from the overall profit and income values, this issue gets even more accentuated with more realistic weather models (i.e., $W1$).
The weather stochasticity is reflected on the produced power, which causes higher reimbursement and supply costs to fulfill the SLAs.
This demands approaches able to dynamically tune the amount of saleable provisioning agreements.
\section{Modeling supply and demand to improve profit}\label{sec:annevaluation}
\subsection{Broker modeling and training}
As seen, to effectively seek profit maximization the power broker needs to adapt its behavior to the expected sequence of events going to occurr in the microgrid.
For each neuron $i$, the learning process consists in the optimization of the weights $w_{ji}$ associated to the incoming connection from each neuron $j$, as well as the bias $b_{i}$.
The broker's input layer includes $P_{re}$, $fit$, $P_{grid}$ and $get$.
The output of a neuron $i$ can be computed for a generic step k using the activation function as:
\begin{equation}\label{eq:feed_forward}
o_{i}(k) = F(\sum_{j=0}^{n} w_{ji}o_{j}(k-1)+b_{i})
\end{equation}
In this study, F is a simple linear threshold function:
\begin{equation}\label{eq:linear_activation}
    F(x)=
    \begin{cases}
      -1.0, & \text{if} x \leq -1.0,\\
      x, & \textit{if}\ -1.0 < x < 1.0,\\
      1.0, & \text{otherwise}
    \end{cases}
\end{equation}
The output layer includes a price for each provisioning duration, namely for the unitary agreement, as well as for the SLAs lasting respectively 10, 30, 60, 120, 600 and 1800 seconds.

\begin{figure}[h!]
	\centering
		\includegraphics[trim=12.82cm 23.82cm 4.75cm 4cm,clip,width=0.4\columnwidth]{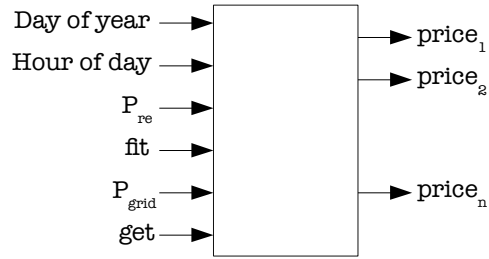} 
	\caption{Proposed ANN interface}\label{fig:ann_a}
\end{figure}

The first design criteria concerns the representation of contextual information, which in presence of photovoltaics is related to sunlight availability.
One possibility to model seasonal patterns is to use i) an input for the hour of the day and ii) one for the day of the year (Fig.~\ref{fig:ann_a}).
This can be modeled using gaussian or sinusoidal functions. In particular, we used a simple model to reflect the higher availability of light in the central part of the day: $\sin( \pi \cdot \frac{t }{ t_{max} } )$, with t used to indicate either the hour of the day or the day of the year.
A neural network was trained using evolutionary algorithms (See Table~\ref{tab:evoparameters}), for each of the scenario previously used to assess the rule-based brokers.
This includes: ideal and real weather conditions, different season and different grid provisioning plan.

In particular, the networks were trained on 1 day simulation data (see Fig.~\ref{fig:weather}) at 1 Hz resolution.
The evolution is driven by the economic profit, as in Eq.~\ref{eq:fitness}.
We initially limited the training to 800 generations.
However, since we noticed stabilization of the fitness already before 500 generations we shortened the simulation for time issues.
Given the predefined $fit$ and $get$ price models, the broker can seek profit maximization by modeling the expectation of future resource availability.
We therefore further penalize the reimbursement cost $C_{reimbursement}$ by multiplying it to a reimbursement penalty $\delta_{r}$ which we empirically set to 100000.
\begin{table}[h!t]
 \centering
 \caption{Parameters of the evolutionary algorithm}\label{tab:evoparameters}
	\begin{tabular}{l r}
    \hline
    Population size					& 50\\
    Number of generations			& 500\\
    Elite selection rate [\%]		& 15\\
    Mutation rate [\%]				& 40\\
    Crossover rate [\%]				& 30\\
    Random-creation rate [\%]		& 5\\
    Random-selection rate [\%]		& 10\\
    \hline
    \end{tabular}
\end{table}
While artificial neural networks are universal function approximators, their ability to learn a function is in fact greatly affected by their topology.
The number of neurons in the hidden layer affects the ability to generalize their experience, leading to overfitting when using fewer neurons and underfitting when using too many.
The optimal number of hidden neurons depends on the complexity of the function to be approximated, and, therefore, indirectly on the number of input and output nodes~\cite{zhevzhyk:15}. 
There exist empirically derived rules-of-thumb for selecting the number of hidden neurons providing a range of possible configurations~\cite{Blum92,Swingler96,Berry97,Boger97,Caudill93}.
In this work we selected the number of hidden neurons based on a experiments within the suggested range of 2,3, and 4 neurons, where 2 neurons showed best behavior.
Moreover, we applied two different representations: three-layered ANNs and fully connected ANNs.
This allows for the assessment of both simple feedforward and more expressive recurrent structures.
\subsection{Brokers for specific scenarios}
\begin{figure}[h!]
	\centering
	\begin{subfigure}[b]{0.85\columnwidth}
 		\includegraphics[trim=0.2cm 0cm 0.28cm 0cm,clip,width=0.98\columnwidth]{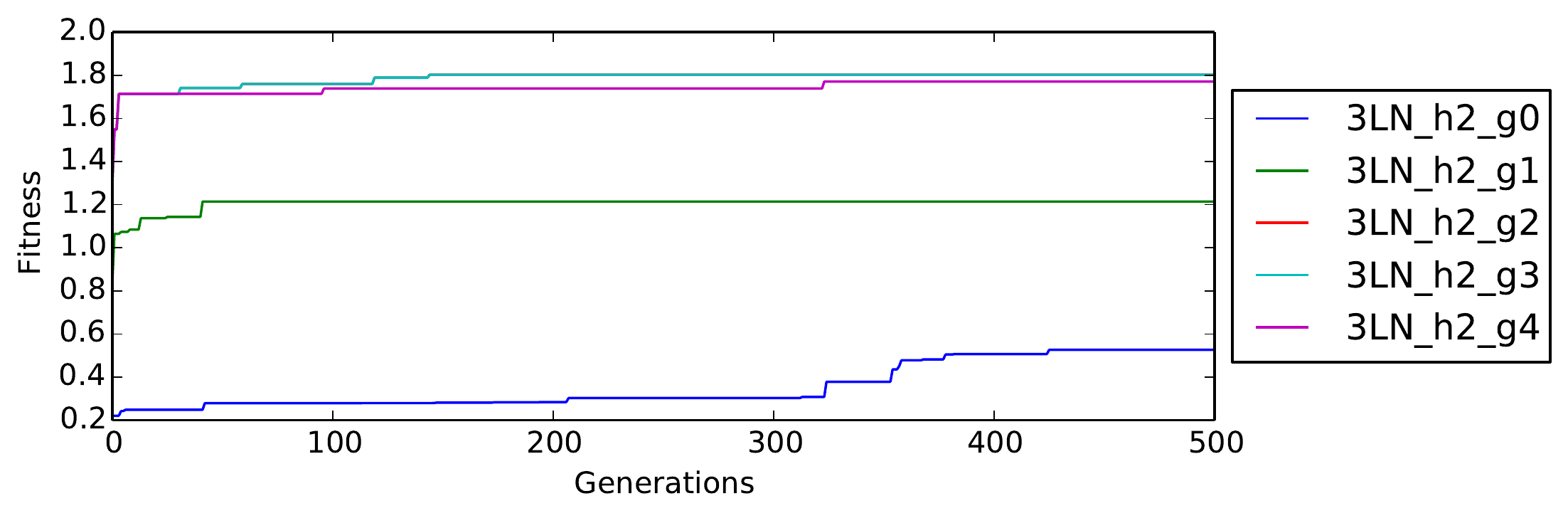} 
 		\caption{Training on the winter day}\label{fig:ann_a:fitness_result:winter}
 	\end{subfigure}%
	
	\begin{subfigure}[b]{0.85\columnwidth}
 		\includegraphics[trim=0.2cm 0cm 0.28cm 0cm,clip,width=0.98\columnwidth]{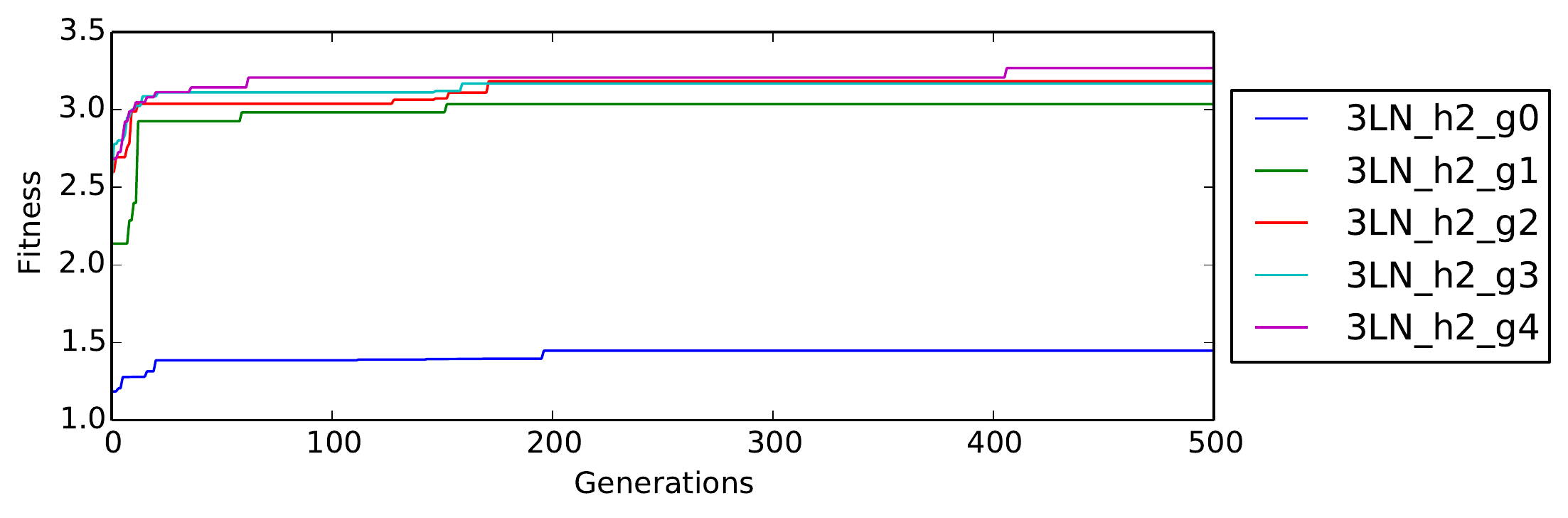} 
 		\caption{Training on the summer day}\label{fig:ann_a:fitness_result:summer}
 	\end{subfigure}%
	
	\caption{Fitness landscape for the proposed model}\label{fig:ann_a:fitness_result}
\end{figure}

\begin{figure}
	\centering
	\begin{subfigure}[b]{0.7\columnwidth}
		\includegraphics[width=\columnwidth]{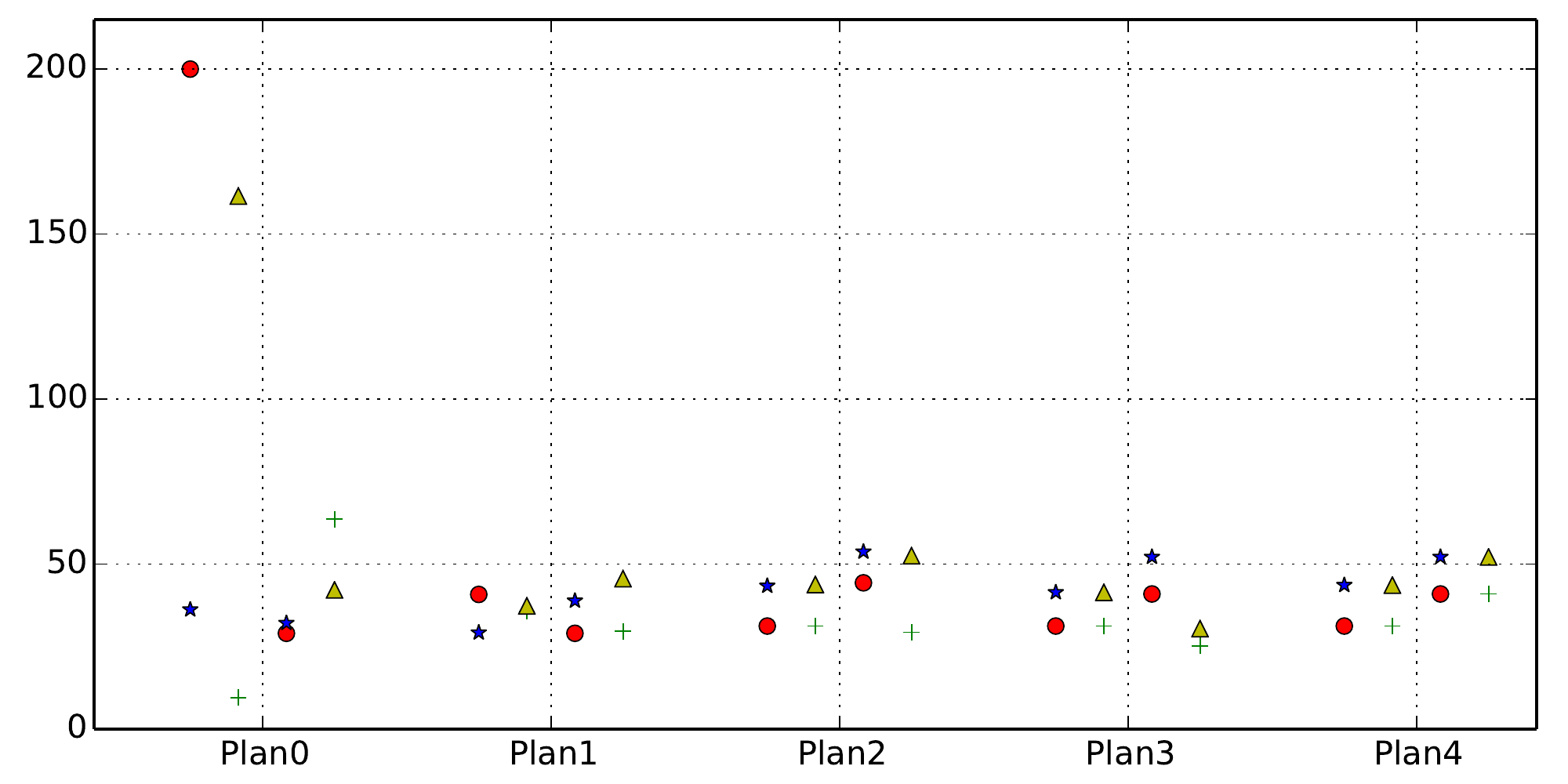}
		\caption{PAR for different $P_{grid}$ scenarios}\label{fig:evaluation_ann_a_1:par}
	\end{subfigure}
	~
	\begin{subfigure}[b]{0.7\columnwidth}
		\includegraphics[ width=\columnwidth]{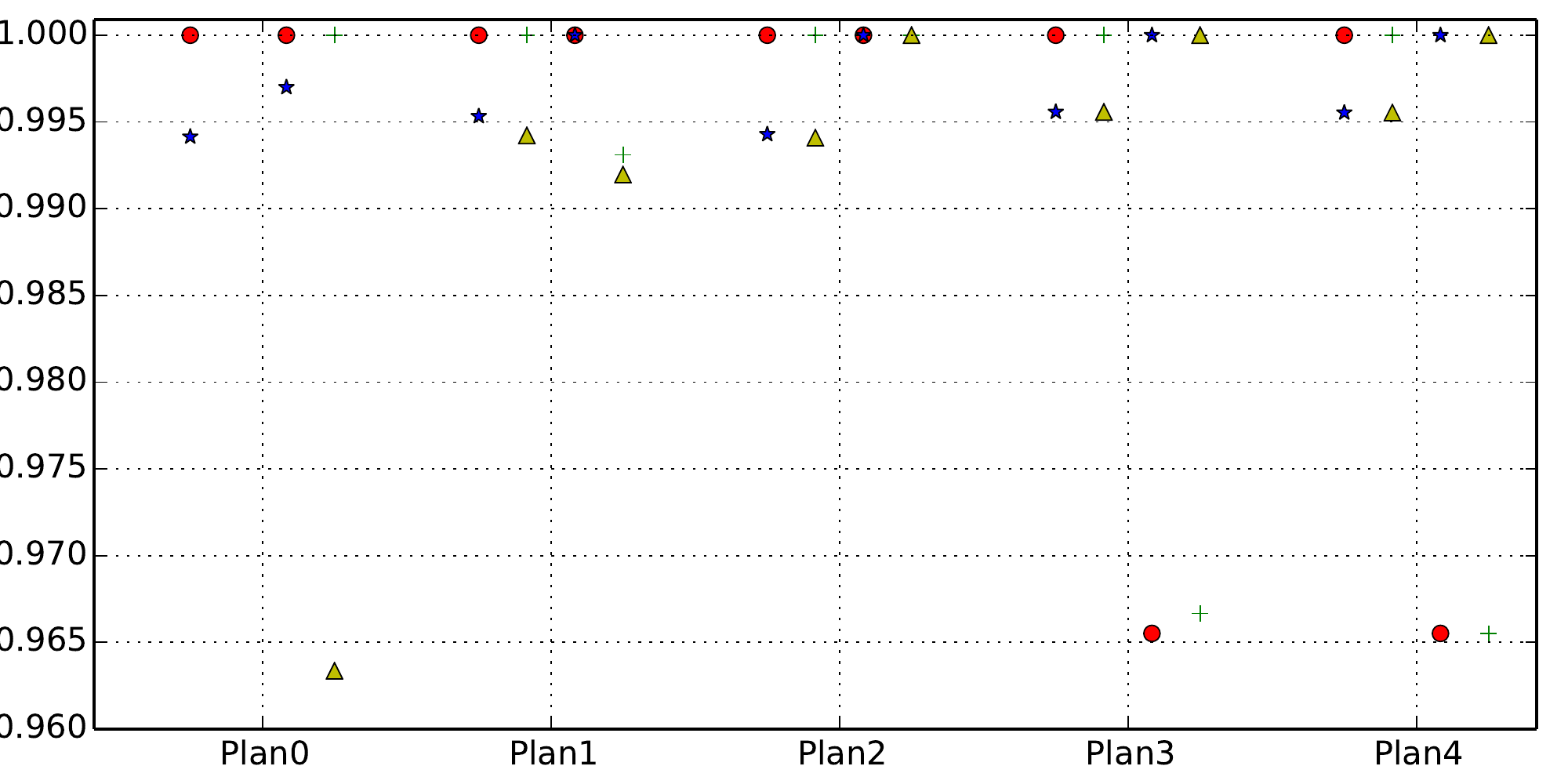}
		\caption{Service availability for different $P_{grid}$ scenarios}
	\end{subfigure}
	~
	\begin{subfigure}[b]{0.7\columnwidth}
		\includegraphics[width=\columnwidth]{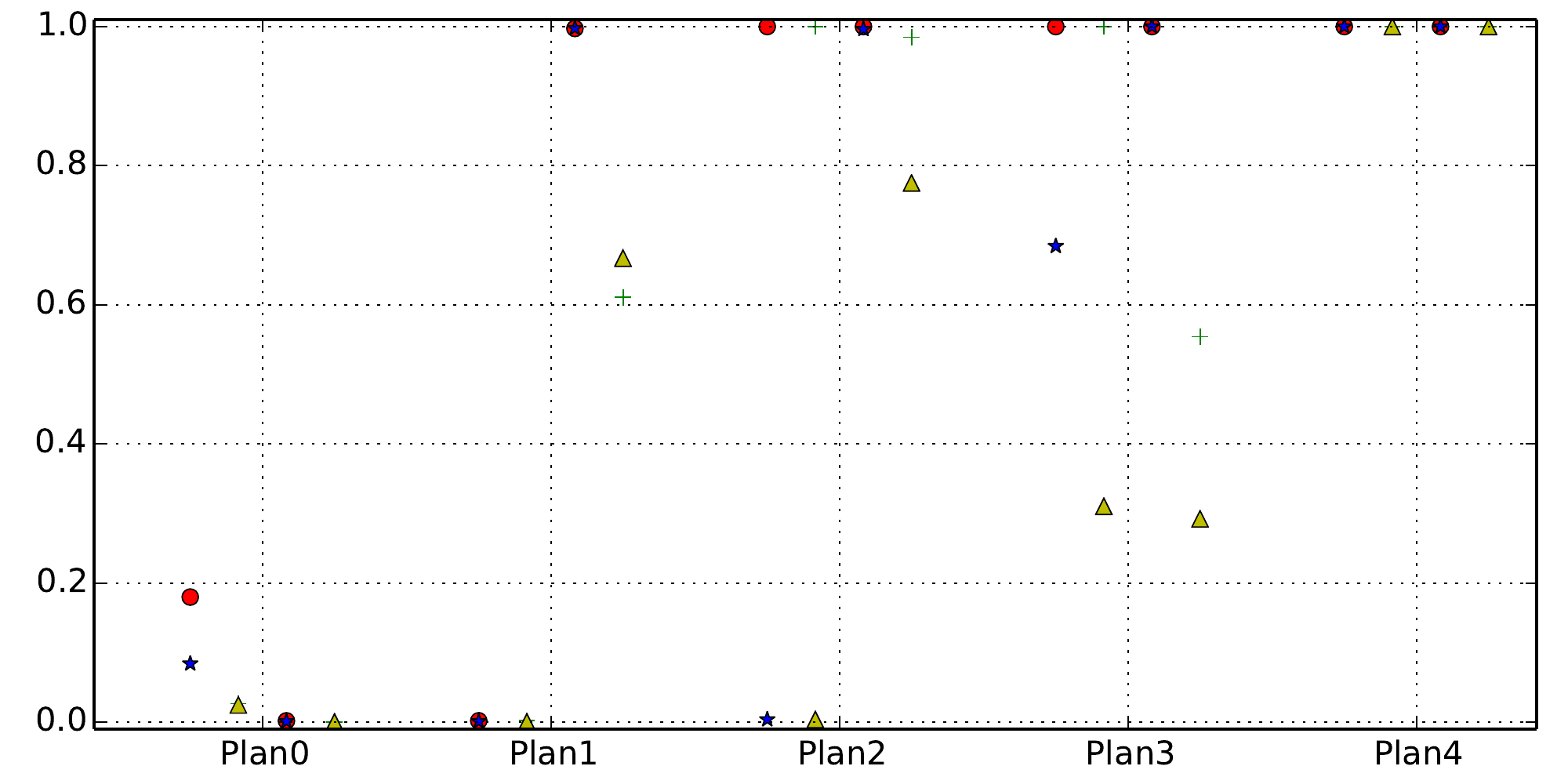}
		\caption{System reactivity for different $P_{grid}$ scenarios}
	\end{subfigure}
	~
	\begin{subfigure}[b]{0.7\columnwidth}
		\includegraphics[ width=\columnwidth]{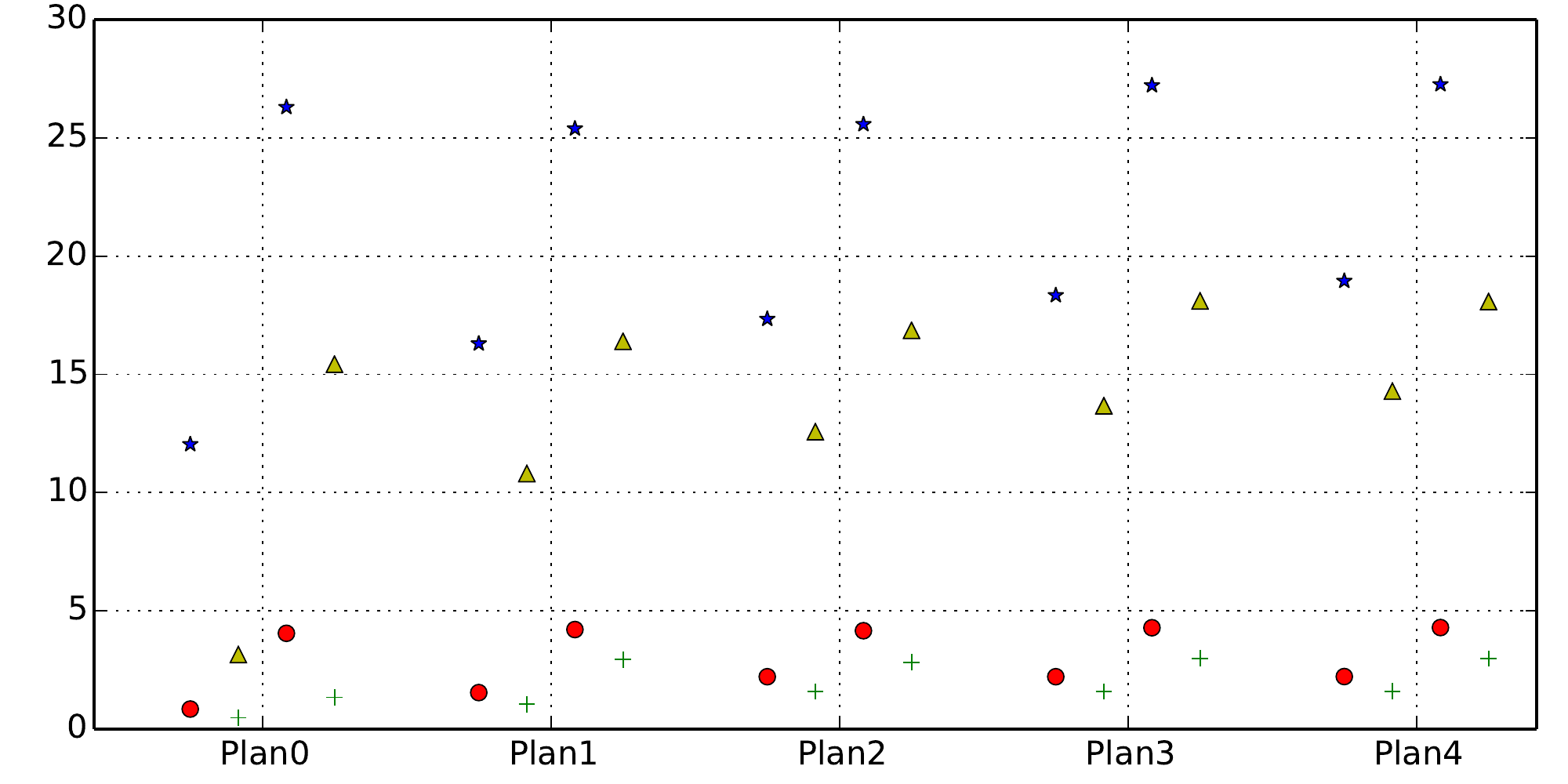}
		\caption{Broker's profit for different $P_{grid}$ scenarios}
	\end{subfigure}
	\caption{Evaluation metrics for the proposed broker's model}\label{fig:evaluation_ann_a_1}
\end{figure}

\begin{figure}
	\centering
	\begin{subfigure}[b]{0.7\columnwidth}
		\includegraphics[ width=\columnwidth]{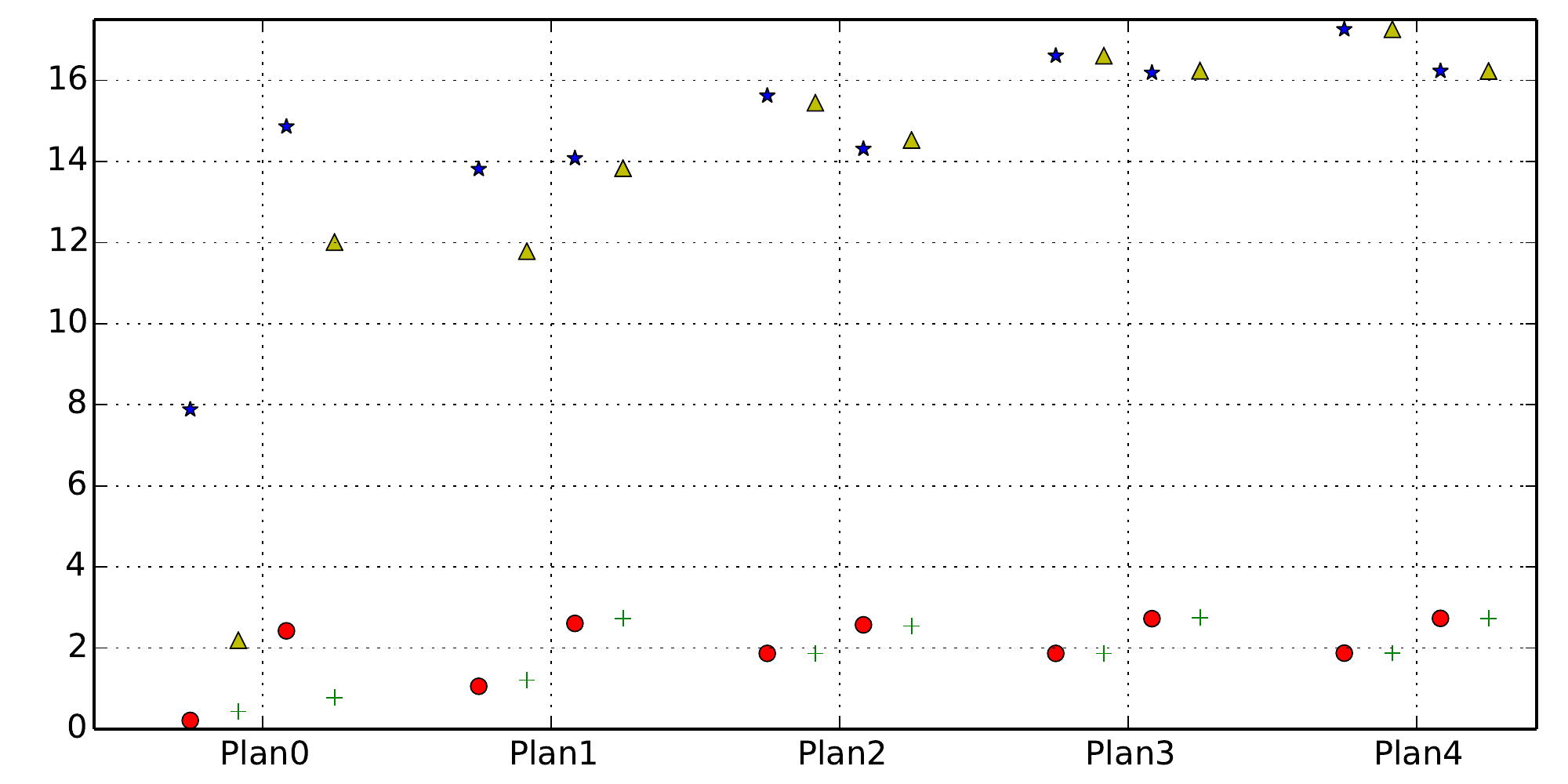}
		\caption{Income from local sale for different $P_{grid}$ scenarios}
	\end{subfigure}
	~
	\begin{subfigure}[b]{0.7\columnwidth}
		\includegraphics[ width=\columnwidth]{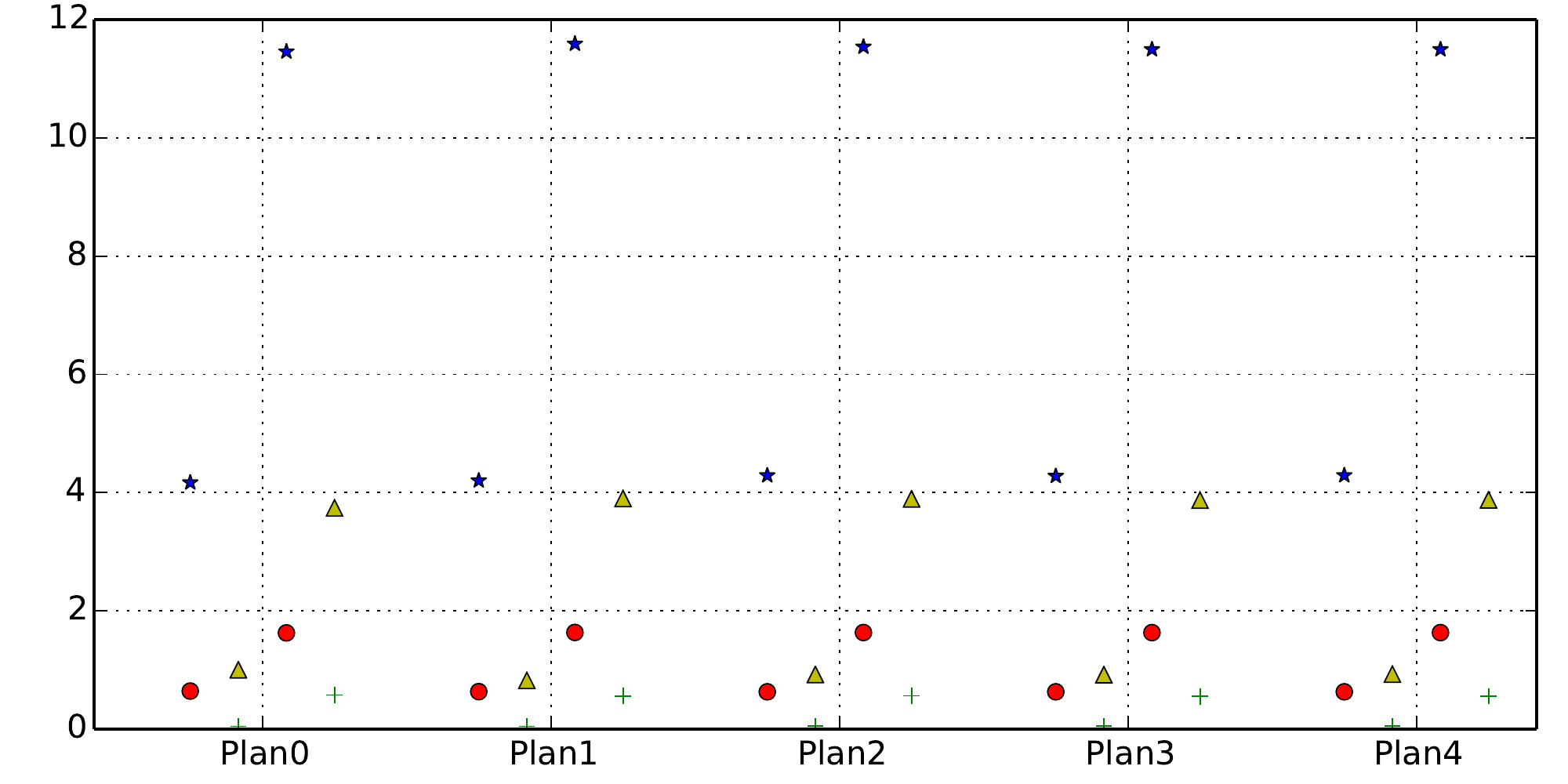}
		\caption{Income from feed-in for different $P_{grid}$ scenarios}
	\end{subfigure}
	~
	\begin{subfigure}[b]{0.7\columnwidth}
		\includegraphics[width=\columnwidth]{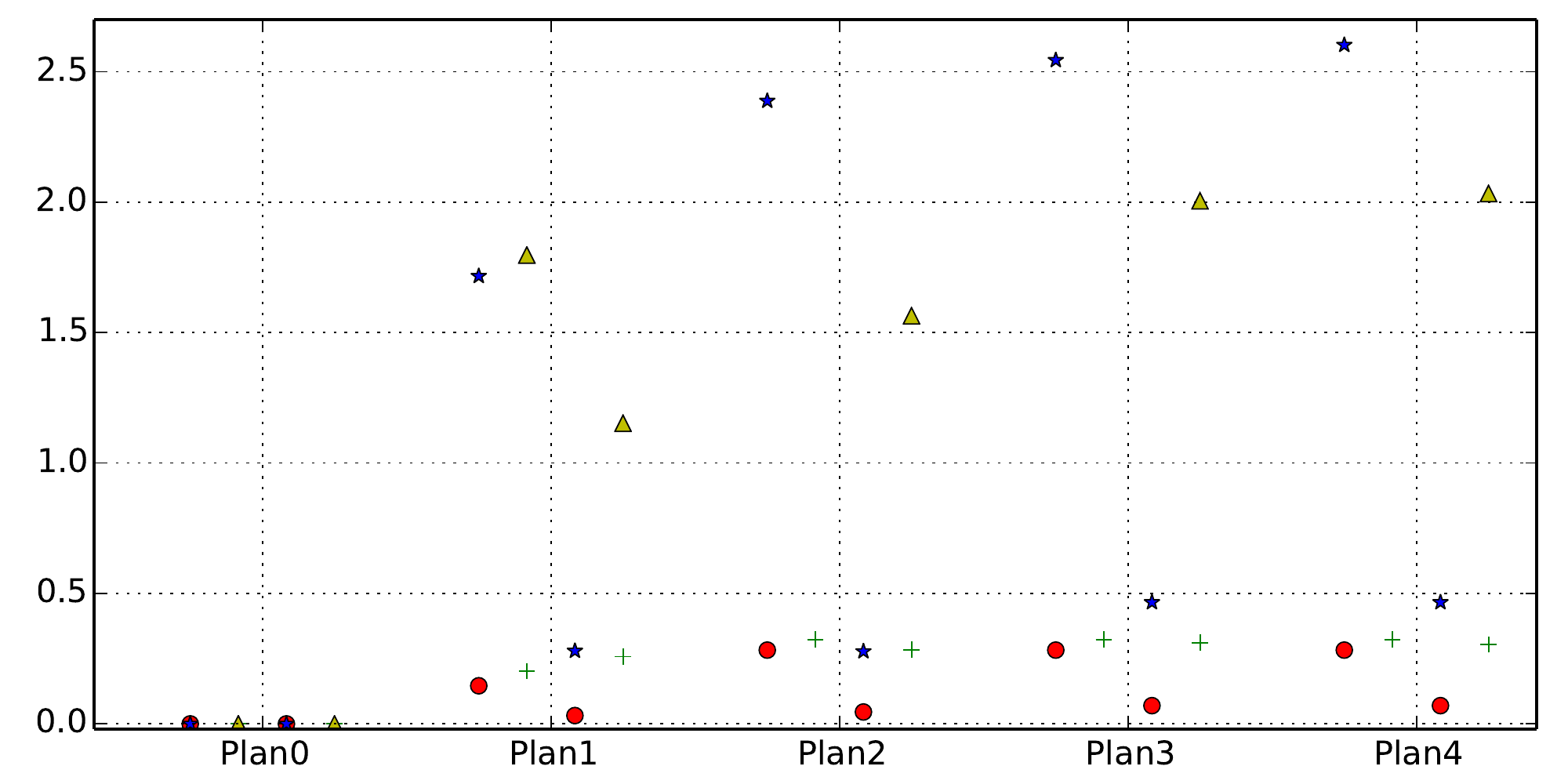}
		\caption{Supply costs for different $P_{grid}$ scenarios}
	\end{subfigure}
	~
	\begin{subfigure}[b]{0.7\columnwidth}
		\includegraphics[width=\columnwidth]{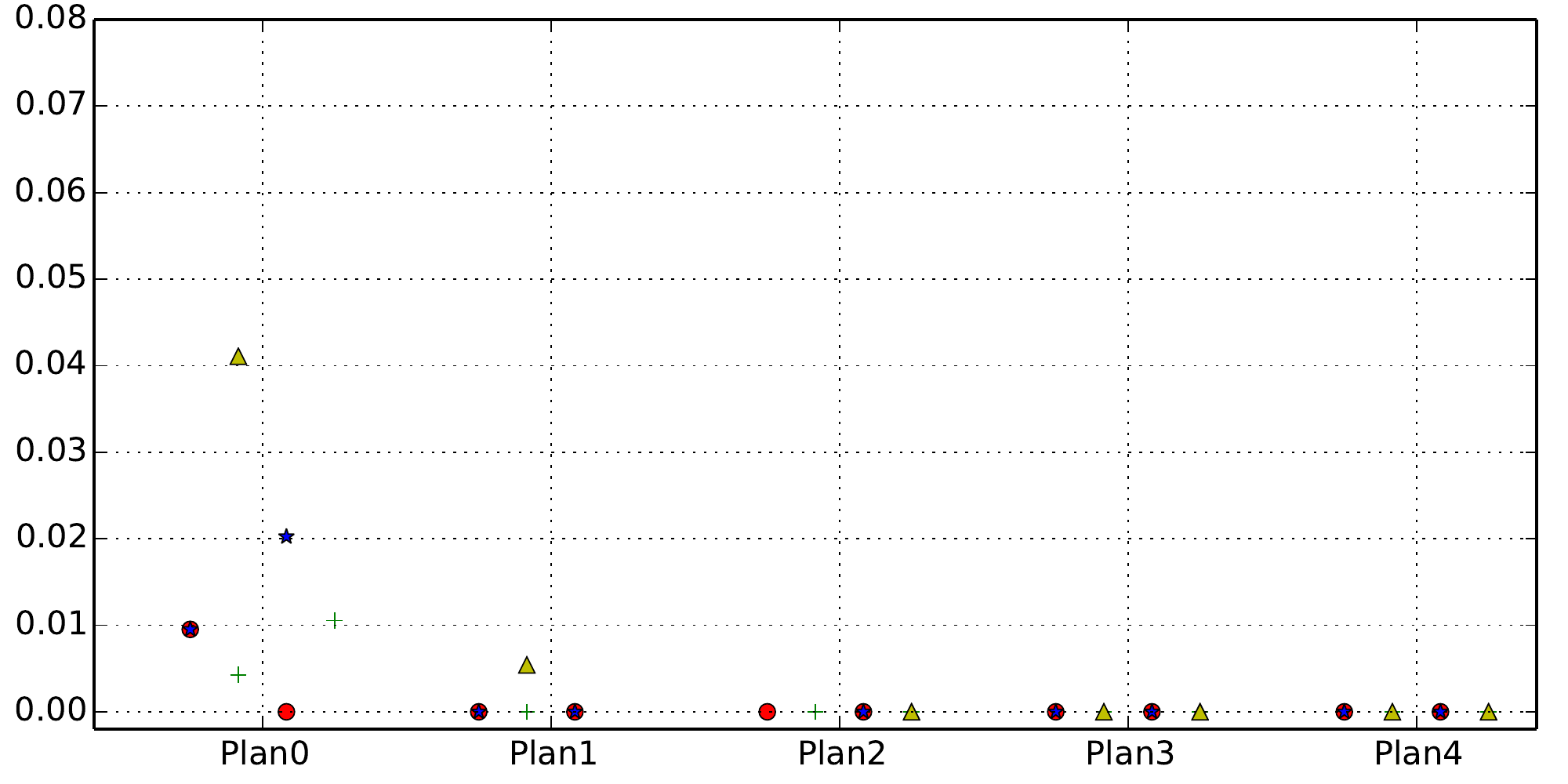}
		\caption{Reimbursement costs for different $P_{grid}$ scenarios}\label{fig:evaluation_ann_a_2:reimbursement}
	\end{subfigure}
	\caption{Profit components the proposed broker's model}\label{fig:evaluation_ann_a_2}
\end{figure}

In the first experiment, we assess the fitness landscape over different scenarios, namely:
i) different grid energy provisioning plans and ii) different season and weather conditions.
Fig.~\ref{fig:ann_a:fitness_result} shows the fitness landscape for the proposed model over the winter (Fig.~\ref{fig:ann_a:fitness_result:winter}) and summer season (Fig.~\ref{fig:ann_a:fitness_result:summer}).
Differences in the availability of both renewable and grid-supplied energy determine different fitness for the broker.
%
%
%
%
%
The experiment continues by evaluating the different brokerage models in each training scenario, using the selected performance measures.
This allows for a comparison of the learned brokers against the rule-based brokers.
Accordingly, we selected the best candidate network (i.e., in the last generation) for use over respectively 1 day and 1 week time.
Moreover, the learned brokers were placed under both ideal and real weather conditions (i.e., as in their simulation enviroment).
Fig.~\ref{fig:evaluation_ann_a_1} show the main performance metrics.
For each grid plan the points on the left and on the right to the label represent respectively results for the winter and the summer season.
Moreover, the symbols correspond to: i) ideal weather for 1 day (circle), ii) ideal weather for 1 week (star), iii) real weather for 1 day (plus) and iv) real weather for 1 week (triangle).
As for the rule-based brokers, higher amount of power resulting from the main grid or renewable sources allows for higher market volume and consequently profit.
Accordingly, in Fig.~\ref{fig:evaluation_ann_a_1:par} the high PAR for the $Plan_{0}$ is due to the absence of $P_{grid}$,
which makes not possible the operation of certain loads and results in $P_{max} = 2 kW$ and $P_{avg} = 10 W$.
Similar differences are encountered for the summer weather as opposed to the winter weather, as well as the ideal weather with respect to the actual weather.
A further remark is that the brokers correctly learned to minimize the reimbursement costs, as compared to their rule-based counterparts (see Fig.~\ref{fig:evaluation_ann_a_2:reimbursement}).
To observe the different behavior of learned brokers for the different scenarios
we report in Table~\ref{tab:SLA_duration_evolution} the number of sold provisioning durations, respectively for 1 day and 7 days long simulation.
The first part reports the results of the winter weather, whilst the second for the summer.
Given the aggressive trading attitude of designed loads (i.e., independent from their price sensitivity),
the brokers' pricing mechanism has more evident effects in the resource constrained settings, as in $Plan_{0}$ and $Plan_{1}$.
Therein the effects of real weather conditions significantly affect the global power availability, which makes shorter SLAs favorite.
Contrarily, in other provisioning plans even with very stochastic weather the power supplied by the main grid is normally enough to back the loads.
Effects can therefore be observed on an economical basis, with the broker increasing the price for the resource proportionally to the grid energy tariff.
\begin{table*}[h!t]
 \centering
 \caption{Traded service-level agreements}\label{tab:SLA_duration_evolution}
	\begin{tabular}{| l l | l l l l l l l |}
	\hline
		Plan				&	Weather			&	\multicolumn{7}{c}{SLA duration}\\
    \hline
    						&					&	0	&	10	&	30		&	60		&	120		&	600		&	1800\\
    \hline
    \multirow{ 2}{*}{0}		&	Ideal			& 0/0	&	0/0	&	8/51	&	1/24	&	0/98	&	8/89	&	0/3\\
    						&	Real			& 0/0	&	0/1 &	5/41	&	0/3		&	0/5		&	11/55	&	2/5\\
    \hline
    \multirow{ 2}{*}{1}		&	Ideal			& 0/0	&	0/0	&	14/156	&	2/34	&	0/104	&	38/290	&	0/0\\
    						&	Real			& 0/0	&	0/0	&	21/152	&	1/23	&	0/64	&	45/276	&	0/0\\
    \hline
    \multirow{ 2}{*}{2}		&	Ideal			& 0/0	&	0/0	&	21/144	&	3/43	&	2/106	&	47/280	&	0/0\\
    						&	Real			& 0/0	&	0/0	&	21/137	&	3/43	&	2/106	&	47/273	&	0/0\\
    \hline
    \multirow{ 2}{*}{3}		&	Ideal			& 0/0	&	0/0	&	21/159	&	3/43	&	2/108	&	47/297	&	0/0\\
    						&	Real			& 0/0	&	0/0	&	21/159	&	3/43	&	2/108	&	47/297	&	0/0\\						
    \hline
    \multirow{ 2}{*}{4}		&	Ideal			& 0/0	&	0/0	&	21/157	&	3/43	&	2/108	&	47/295	&	0/0\\
    						&	Real			& 0/0 	&	0/0	&	21/157	&	3/43	&	2/108	&	47/295	&	0/0\\

    \hline \hline
   	\multirow{ 2}{*}{0}		&	Ideal			& 0/0	&	0/0	&	15/99	&	6/42	&	16/66	&	27/246	&	0/0\\
    						&	Real			& 0/0 	&	0/1	&	15/72	&	2/22	&	2/53	&	23/202	&	0/0\\
    \hline
    \multirow{ 2}{*}{1}		&	Ideal			& 0/0	&	0/0	&	22/158	&	6/42	&	16/66	&	28/184	&	0/0\\
    						&	Real			& 0/0 	&	0/1	&	21/145	&	6/36	&	18/70	&	27/170	&	0/0\\
    	
    \hline
    \multirow{ 2}{*}{2}		&	Ideal			& 0/0	&	0/0	&	19/139	&	6/42	&	16/66	&	25/171	&	2/18\\
    						&	Real			& 0/0 	&	0/0	&	21/156	&	6/42	&	16/66	&	27/200	&	0/6\\
    \hline
    \multirow{ 2}{*}{3}		&	Ideal			& 0/0	&	0/0	&	20/150	&	6/42	&	16/66	&	38/296	&	0/0\\
    						&	Real			& 0/0 	&	0/0	&	21/152	&	6/42	&	16/66	&	39/298	&	0/0\\
    \hline
    \multirow{ 2}{*}{4}		&	Ideal			& 0/0	&	0/0	&	20/150	&	6/42	&	16/66	&	38/296	&	0/0\\
    						&	Real			& 0/0 	&	0/0	&	20/150	&	6/42	&	16/66	&	38/296	&	0/0\\
    \hline
    \end{tabular}
\end{table*}
Consequently, the pricing of SLAs depends strictly on the expectation of future demand.
By setting the price sensitivity to 0.9 (€ / kWh) we simulated the worst possible congestion scenario,
in which depending on the employed usage model all loads desire to operate regardless of the SLA pricing.
In fact, users will assign different price sensitivity models to the loads, according to the delivered utility.
This has the favourable effect of determining an ordering over the loads, and consequently more favourable conditions for the broker and the resulting SLA prices.
%
%
%
%
%
%
%
%
%
\section{Conclusions and future work}\label{sec:conclusions}
Power management in microgrids is an important problem because of the limitations and volatility of resources involved.
A broker is an agent controlling the energy exchanged with and throughout the local power system.
The power broker can be used by both humans and autonomous load controllers to improve operation planning in microgrids.
As opposed to centralised scheduling tools, the broker takes care of price formulation and let the loads to autonomously decide whether to operate.
Consequently, the communication complexity is limited to broadcasting the price across the microgrid.
A power broker is an effective coordinator for the microgrid demand, as:
i) it prevents loads from running when not enough power is available, consequently reducing the overall PAR,
ii) it prices power based based on the proportion produced from renewable sources $P_{re}$,
and iii) it prices provisioning durations to optimize service availability (i.e., optimistic broker) or system reactivity (i.e., pessimistic broker).
Simulations were performed to assess the performance of rule-based brokers on actual demand previously measured in an Austrian household.
We showed the pessimistic and the optimistic brokers respectively yielding service interruption and resource monopolization, as well as economic losses.
This demands solutions able to consider the uncertainty of future supply and demand on the price of provisioning agreements.
This study showed the possibility of exploiting a model of demand and supply to minimize service interruption and operational costs.
In particular, the broker is modeled as an artificial neural network which is progressively evolved based on its gathered economic profit.
Results showed the brokers correctly minimizing reimbursement costs.
In particular, tests were performed against consecutive days previously unused for the training phase and findings were assessed with respect to the initial rule-based solutions.
%
%
%
%
While we showed clear advantages from this approach, different architectures and interfaces should be assessed in further studies.
Moreover, the designer should be released from the burden of designing both the architecture and the sensory input.
A possibility is to evolve the neural architecture beside the synaptic weights, as in the neuroevolution of augmenting topologies (NEAT) \cite{stanley:ec02}.
This would allow for a form of phenotypic plasticity, which would give the broker the possibility to also optimize its topology.
Another aspect to be considered is the possibility to increase the number of hidden layers, as in the so called deep learning.
While this would complicate the learning process, it might allow the broker for learning more sophisticated abstractions over its sensory input.
In conclusion, this study undertook a production-centric vision of the microgrid power management problem.
Whilst the selected approach can learn from the encountered demand, further studies should also assess the actual discomfort caused by the resulting control strategies.
Essentially, those control strategies depend on both employed market mechanisms and strategic-bidding agents.
Therefore, we envisage a full integration of the brokerage tool with the HEMS energy market simulator,
so as to integrate the training of both appliance controllers and power brokers.

\bibliographystyle{alpha}      
\bibliography{docbib.bib}

\end{document}